%% file: main.tex
\documentclass[11pt]{article}
\usepackage[preprint]{acl}
\usepackage{times}
\usepackage{latexsym}
\usepackage[T1]{fontenc}
\usepackage[utf8]{inputenc}
\usepackage{microtype}
\usepackage{amsmath,amssymb,mathtools}
\usepackage{booktabs,array,multirow}
\usepackage{graphicx}
\usepackage{float}
\usepackage{placeins}
\newfloat{algorithm}{tbp}{loa}
\floatname{algorithm}{Algorithm}
\usepackage{tikz}
\usepackage{pgfplots}
\pgfplotsset{compat=1.18}
\usetikzlibrary{arrows.meta,positioning,fit}
\usepackage{xcolor}
\usepackage{url}
\input{macros}

\title{Look Back, Draft Ahead: Corrective Prefix Attention for Parallel Speculative Decoding}
\author{
  \textbf{Mohammad Mahdi Moradi}\thanks{These authors contributed equally to this work.}
  \quad
  \textbf{Ali Pourranjbar}\footnotemark[1] \\
  \textbf{Zongzuo Wang}
  \quad
  \textbf{Mingming Zhu}
  \quad
  \textbf{Ben Feng}
  \quad
  \textbf{Boxing Chen}\thanks{Corresponding author.} \\
  Ascend Team \\
  Toronto Research Center \\
  Huawei Technologies
}

\begin{document}
\maketitle
\begin{abstract}
Parallel speculative decoders propose an entire token block at once, yet one early inconsistency can cause the target to reject its remaining suffix. We introduce Corrective Prefix Attention (CPA), which combines prefix-aware selection with autoregressive supervision to improve these parallel drafts. CPA lets each candidate attend to earlier choices within the block. Fixed candidate shortlists allow all pairwise interactions to be computed before selection, reducing the sequential walk to table lookups, accumulation and lightweight scoring rather than repeated neural evaluations. An auxiliary autoregressive loss trains the shared backbone on clean prefixes without adding an inference pass. Across eight math, code and dialogue benchmarks with Qwen3-4B and Qwen3-8B targets on Ascend NPUs, CPA commits 13--15\% more tokens per verification round than DSpark and 25--26\% more than DFlash. Its mean speedups reach $3.35\times$ and $3.37\times$ under greedy decoding and $2.86\times$ under sampling for both targets. Ablations support both components, while capacity sweeps demonstrate that greater acceptance alone does not guarantee faster decoding. These results motivate correcting parallel proposals with causal information while keeping sequential computation inexpensive. Our implementation is available at
\url{https://github.com/MohammadMahdi1375/vLLM_NPU}.
\end{abstract}

\input{sections/01_introduction}
\input{sections/02_related}
\input{sections/03_method}
\input{sections/04_experiments}

\input{sections/05_results}
\input{sections/06_conclusion}
\FloatBarrier
\label{maintextend}
\input{sections/07_limitations}
\bibliography{references}
\clearpage
\appendix
\input{sections/appendix}
\end{document}

%% file: macros.tex
\newcommand{\result}[1]{\textcolor{blue!65!black}{\textsf{TBD}}}

\newcommand{\ind}{\mathbb{I}}
\newcommand{\R}{\mathbb{R}}
\newcommand{\clip}{\operatorname{clip}}
\newcommand{\argmax}{\operatorname*{arg\,max}}
\newcommand{\softmax}{\operatorname{softmax}}
\newcommand{\TopK}{\operatorname{TopK}}
\newcommand{\CPA}{\textsc{CPA}}
\newcommand{\smalltable}{\small\setlength{\tabcolsep}{4pt}}

%% file: sections/01_introduction.tex
\section{Introduction}
Speculative decoding speeds up a large language model by letting a cheap drafter propose tokens that the target model then checks in a single forward pass \citep{leviathan2023speculative,chen2023speculative}. Autoregressive drafters such as EAGLE-3 run once per proposed token \citep{li2025eagle3}, so long drafts are expensive. Parallel drafters remove this cost. Building on blockwise parallel decoding \citep{stern2018blockwise}, DFlash conditions a small block model on target features and predicts every position of a block in one pass \citep{chen2026dflash}.

Parallel drafting, however, is at odds with how drafts are verified. The target accepts proposals only up to the first token it disagrees with and discards the rest \citep{leviathan2023speculative}. Speed is therefore governed by where the \emph{first} error occurs, and an early error costs far more than a late one. Yet a parallel drafter scores every position before any position has been chosen.

\input{figures/pipeline}

Consider a block in which the candidates are \emph{they} or \emph{she} at one position, \emph{often} at the next, and \emph{run} or \emph{runs} at the third. The correct verb depends on the subject two positions back. A position-wise argmax cannot react to the subject it actually chose. A correction that looks only at the previous token sees \emph{often} on both paths and cannot tell them apart. An autoregressive drafter sees the whole prefix, but pays a backbone step for every token.

Recent work therefore adds a correction stage after the parallel pass. DFlash~2 rescores adjacent candidate pairs \citep{inco2026dflash2}; Domino and DSpark run a recurrent or Markov head through the block \citep{huang2026domino,cheng2026dspark}; D-Loop conditions another backbone pass on the selected prefix \citep{chen2026dloop}. These designs trade context against sequential computation: local rescoring is cheap but sees one token, while recurrent heads and repeated passes see more of the prefix at the price of neural computation at every position.

We show that attention-style correction does not need this trade-off. After the parallel pass, each position has a fixed shortlist of candidates. The interaction between a candidate at position $i$ and a candidate at an earlier position $j$ then depends only on those two tokens and on block features that are already computed, not on which path is eventually chosen. All such interactions can be computed together in one batched step. Attention over the selected prefix then reduces exactly to gathering and accumulating table entries as choices are made, and the sequential walk contains no neural network evaluation.

\emph{Corrective Prefix Attention} (\CPA) builds on this observation and uses prefix information once more during training (Figure~\ref{fig:pipeline}). A parallel drafter only ever learns from masked blocks, so its backbone never practices predicting the next token from a correct prefix, which is exactly the situation that verification rewards. \CPA{} runs the shared backbone a second time on clean, causally masked prefixes and adds a next-token loss over the full vocabulary. This pass exists only during training; inference still runs a single parallel pass.

Our contributions are as follows:
\begin{itemize}\setlength\itemsep{1pt}
\item A selector that attends over all earlier choices in a drafted block, compiled exactly into precomputed candidate-pair tables so that the sequential walk performs only lookups and additions (Section~\ref{sec:method}).
\item An auxiliary autoregressive objective that trains the shared parallel backbone on clean prefixes at no inference cost.
\item An evaluation on eight benchmarks with Qwen3-4B and Qwen3-8B targets on Ascend 910B2 NPUs, in which \CPA{} has the highest mean speedup at both temperatures and commits 13--15\% more tokens per round than DSpark. Attending to older choices raises acceptance and speed by about 5\% over a local-only selector with similar inferred round cost, the autoregressive loss improves both on all eight benchmarks, and capacity sweeps show when more acceptance stops paying for its cost.
\end{itemize}

%% file: figures/pipeline.tex
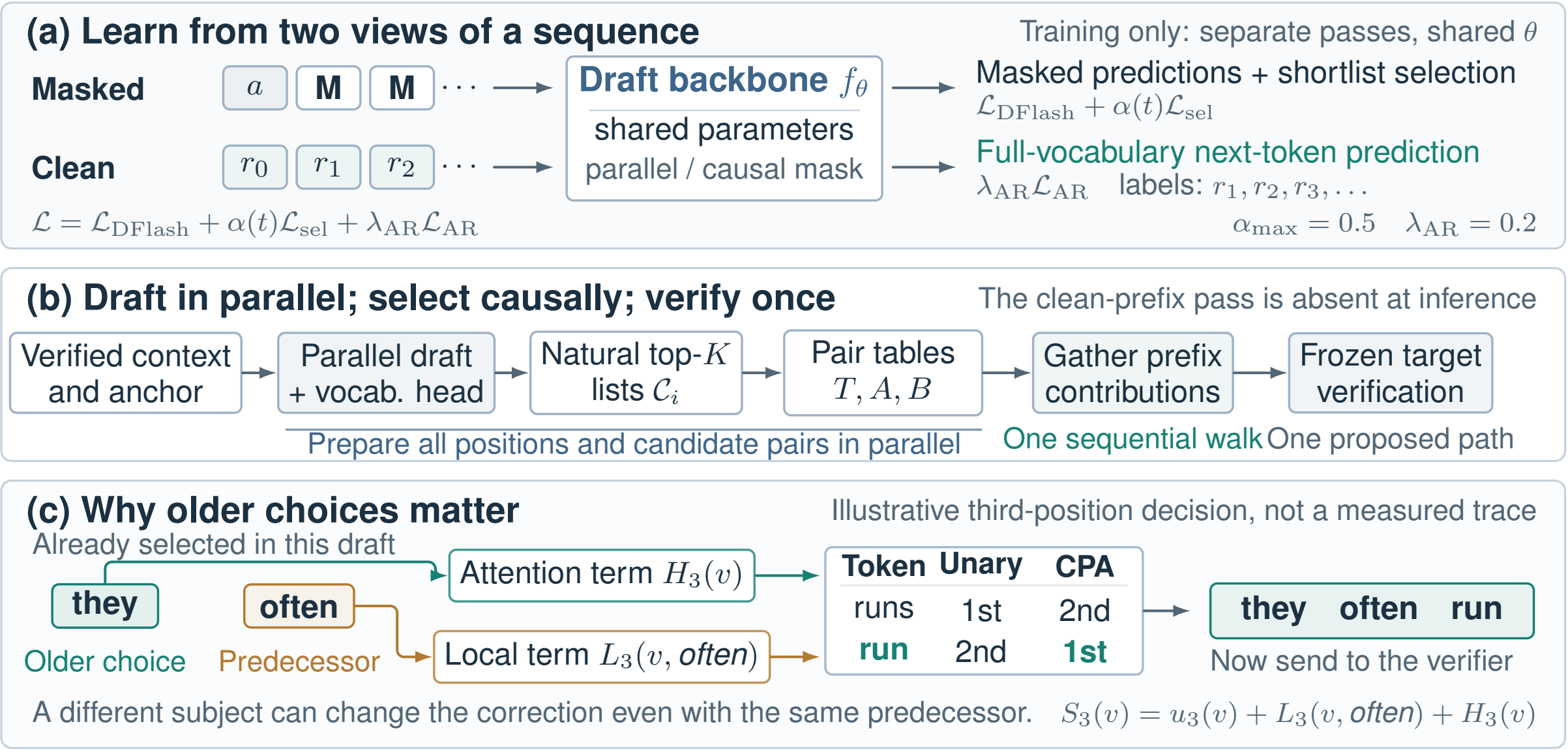
\begin{figure*}[t]
\centering
\resizebox{\textwidth}{!}{\input{figures/architecture}}
\caption{\textbf{Corrective Prefix Attention (CPA).} (a) A clean, causally masked training pass teaches the shared drafter to predict shifted reference tokens over the full vocabulary. The masked pass learns parallel proposals and their shortlist selection; both views use pre-anchor target features. (b) Only the parallel pass runs at inference; local and history pair scores are prepared before the causal selection walk. (c) Attention to an older subject can change the preferred verb even when the immediate predecessor is fixed. M denotes a mask token. The rank changes are schematic; the anchor and other candidates are omitted.}
\label{fig:pipeline}
\end{figure*}

%% file: figures/architecture.tex
% Editable, original schematic. Token strings and rank changes are illustrative.
\begin{tikzpicture}[
 x=1cm,y=1cm,line cap=round,line join=round,
 font=\sffamily\fontsize{9}{10.5}\selectfont,
 txt/.style={anchor=west,inner sep=0pt,align=left,text=cpaInk},
 title/.style={txt,font=\sffamily\bfseries\fontsize{10.5}{12}\selectfont},
 fine/.style={txt,font=\sffamily\fontsize{8.4}{9.8}\selectfont,text=cpaMuted},
 box/.style={draw=cpaBlue!50,fill=white,rounded corners=2pt,line width=.65pt,
             align=center,inner sep=3pt,text=cpaInk},
 token/.style={box,font=\sffamily\bfseries\fontsize{9.4}{11}\selectfont,
               minimum height=.44cm,minimum width=.65cm,inner sep=2pt},
 arrow/.style={-{Latex[length=1.8mm,width=1.25mm]},line width=.75pt,draw=cpaMuted},
 center/.style={inner sep=0pt,align=center,text=cpaInk}]
\definecolor{cpaInk}{HTML}{1B3040}
\definecolor{cpaMuted}{HTML}{526574}
\definecolor{cpaBlue}{HTML}{3E6589}
\definecolor{cpaTeal}{HTML}{168074}
\definecolor{cpaGold}{HTML}{B27932}
\path[use as bounding box] (0,0) rectangle (16,7.65);

% (a) Same parameters, separate activations and attention masks.
\filldraw[rounded corners=3pt,draw=cpaBlue!30,fill=cpaBlue!3,line width=.6pt]
 (.02,5.12) rectangle (15.98,7.63);
\node[title] at (.27,7.31) {(a) Learn from two views of a sequence};
\node[fine,anchor=east] at (15.71,7.31) {Training only: separate passes, shared $\theta$};
\node[txt,font=\sffamily\bfseries\fontsize{9}{11}\selectfont] at (.32,6.75) {Masked};
\node[token,fill=cpaBlue!10] at (2.61,6.76) {$a$};
\node[token] at (3.36,6.76) {M};
\node[token] at (4.11,6.76) {M};
\node[center] at (4.72,6.76) {$\cdots$};
\node[txt,font=\sffamily\bfseries\fontsize{9}{11}\selectfont] at (.32,5.95) {Clean};
\node[token,fill=cpaTeal!8] at (2.61,5.95) {$r_0$};
\node[token,fill=cpaTeal!8] at (3.36,5.95) {$r_1$};
\node[token,fill=cpaTeal!8] at (4.11,5.95) {$r_2$};
\node[center] at (4.72,5.95) {$\cdots$};
\draw[arrow] (5.05,6.76)--(5.66,6.76);
\draw[arrow] (5.05,5.95)--(5.66,5.95);
\filldraw[box,fill=white] (5.79,5.61) rectangle (9.01,7.08);
\node[center,font=\sffamily\bfseries\fontsize{10}{12}\selectfont,text=cpaBlue] at (7.40,6.82) {Draft backbone $f_\theta$};
\draw[cpaBlue!30] (6.02,6.53)--(8.78,6.53);
\node[center] at (7.40,6.31) {shared parameters};
\node[center,font=\sffamily\fontsize{8.4}{10}\selectfont,text=cpaMuted] at (7.40,5.91) {parallel / causal mask};
\draw[arrow] (9.12,6.76)--(9.77,6.76);
\draw[arrow] (9.12,5.95)--(9.77,5.95);
\node[txt] at (9.96,6.89) {Masked predictions + shortlist selection};
\node[fine] at (9.96,6.56) {$\mathcal L_{\mathrm{DFlash}}+\alpha(t)\mathcal L_{\mathrm{sel}}$};
\node[txt,text=cpaTeal] at (9.96,6.08) {Full-vocabulary next-token prediction};
\node[fine] at (9.96,5.75) {$\lambda_{\mathrm{AR}}\mathcal L_{\mathrm{AR}}$\quad labels: $r_1,r_2,r_3,\ldots$};
\node[fine] at (.32,5.37) {$\mathcal L=\mathcal L_{\mathrm{DFlash}}+\alpha(t)\mathcal L_{\mathrm{sel}}+\lambda_{\mathrm{AR}}\mathcal L_{\mathrm{AR}}$};
\node[fine,anchor=east] at (15.70,5.37) {$\alpha_{\max}=0.5$\quad $\lambda_{\mathrm{AR}}=0.2$};

% (b) Serving path. Only one full-vocabulary draft projection is performed.
\filldraw[rounded corners=3pt,draw=cpaBlue!30,fill=white,line width=.6pt]
 (.02,2.95) rectangle (15.98,4.92);
\node[title] at (.27,4.59) {(b) Draft in parallel; select causally; verify once};
\node[fine,anchor=east] at (15.70,4.59) {The clean-prefix pass is absent at inference};
\node[box,minimum width=2.05cm,minimum height=.78cm] (context) at (1.29,3.85) {Verified context\\and anchor};
\node[box,fill=cpaBlue!7,minimum width=2.05cm,minimum height=.78cm] (draft) at (3.95,3.85) {Parallel draft\\+ vocab. head};
\node[box,minimum width=1.80cm,minimum height=.78cm] (lists) at (6.50,3.85) {Natural top-$K$\\lists $\mathcal C_i$};
\node[box,minimum width=2.02cm,minimum height=.78cm] (tables) at (9.02,3.85) {Pair tables\\$T,A,B$};
\node[box,fill=cpaTeal!7,minimum width=2.00cm,minimum height=.78cm] (walk) at (11.57,3.85) {Gather prefix\\contributions};
\node[box,fill=cpaBlue!7,minimum width=1.95cm,minimum height=.78cm] (verify) at (14.20,3.85) {Frozen target\\verification};
\draw[arrow] (context.east)--(draft.west);
\draw[arrow] (draft.east)--(lists.west);
\draw[arrow] (lists.east)--(tables.west);
\draw[arrow] (tables.east)--(walk.west);
\draw[arrow] (walk.east)--(verify.west);
\draw[cpaBlue!65,line width=.65pt] (2.93,3.27)--(10.02,3.27);
\node[center,font=\sffamily\fontsize{8.4}{10}\selectfont,text=cpaBlue] at (6.48,3.10) {Prepare all positions and candidate pairs in parallel};
\node[center,font=\sffamily\fontsize{8.4}{10}\selectfont,text=cpaTeal] at (11.57,3.16) {One sequential walk};
\node[center,font=\sffamily\fontsize{8.4}{10}\selectfont,text=cpaMuted] at (14.20,3.16) {One proposed path};

% (c) Concrete schematic for the two distinct sources of prefix information.
\filldraw[rounded corners=3pt,draw=cpaBlue!30,fill=cpaBlue!2,line width=.6pt]
 (.02,.02) rectangle (15.98,2.75);
\node[title] at (.27,2.42) {(c) Why older choices matter};
\node[fine,anchor=east] at (15.70,2.42) {Illustrative third-position decision, not a measured trace};
\node[fine] at (.33,2.08) {Already selected in this draft};
\node[token,draw=cpaTeal,fill=cpaTeal!12,minimum width=1.08cm] (old) at (1.08,1.47) {they};
\node[token,draw=cpaGold,fill=cpaGold!10,minimum width=1.12cm] (prev) at (3.06,1.47) {often};
\node[center,font=\sffamily\fontsize{8.4}{10}\selectfont,text=cpaTeal] at (1.08,.91) {Older choice};
\node[center,font=\sffamily\fontsize{8.4}{10}\selectfont,text=cpaGold] at (3.06,.91) {Predecessor};
\node[box,draw=cpaTeal!80,fill=white,minimum width=2.75cm,minimum height=.50cm] (history) at (6.15,1.78) {Attention term $H_3(v)$};
\node[box,draw=cpaGold!80,fill=white,minimum width=2.75cm,minimum height=.50cm] (local) at (6.15,.95) {Local term $L_3(v,\text{\emph{often}})$};
\draw[arrow,draw=cpaTeal,rounded corners=2pt] (old.north)--(1.08,1.92)--(4.45,1.92)--(4.45,1.78)--(history.west);
\draw[arrow,draw=cpaGold,rounded corners=2pt] (prev.east)--(4.06,1.47)--(4.06,.95)--(local.west);
\node[box,draw=cpaBlue!45,fill=white,minimum width=3.24cm,minimum height=1.30cm] (ranks) at (10.05,1.42) {};
\node[center,font=\sffamily\bfseries\fontsize{8.6}{10}\selectfont] at (9.03,1.88) {Token};
\node[center,font=\sffamily\bfseries\fontsize{8.6}{10}\selectfont] at (10.02,1.88) {Unary};
\node[center,font=\sffamily\bfseries\fontsize{8.6}{10}\selectfont] at (11.08,1.88) {CPA};
\draw[cpaBlue!20] (8.59,1.68)--(11.51,1.68);
\node[center] at (9.03,1.43) {runs};
\node[center] at (10.02,1.43) {1st};
\node[center] at (11.08,1.43) {2nd};
\node[center,text=cpaTeal,font=\sffamily\bfseries\fontsize{9}{10.5}\selectfont] at (9.03,1.01) {run};
\node[center] at (10.02,1.01) {2nd};
\node[center,text=cpaTeal,font=\sffamily\bfseries\fontsize{9}{10.5}\selectfont] at (11.08,1.01) {1st};
\draw[arrow,draw=cpaTeal] (history.east)--(8.38,1.78);
\draw[arrow,draw=cpaGold] (local.east)--(8.38,.95);
\draw[arrow] (ranks.east)--(12.16,1.42);
\node[token,draw=cpaTeal,fill=cpaTeal!10,minimum width=3.30cm,minimum height=.56cm] at (14.01,1.42) {they\quad often\quad run};
\node[fine] at (12.35,.92) {Now send to the verifier};
\node[fine] at (.33,.36) {A different subject can change the correction even with the same predecessor.};
\node[fine,anchor=east] at (15.70,.36) {$S_3(v)=u_3(v)+L_3(v,\text{\emph{often}})+H_3(v)$};
\end{tikzpicture}

%% file: sections/02_related.tex
\section{Related Work}
\paragraph{Drafting for speculative decoding.}
Speculative decoding verifies drafted tokens with the target model and, with rejection sampling, preserves its output distribution \citep{leviathan2023speculative,chen2023speculative}. Autoregressive drafters such as EAGLE-3 reuse target features and produce one token per drafter step \citep{li2025eagle3}. Blockwise drafters instead predict a whole block at once \citep{stern2018blockwise}. DFlash conditions a small block model on target hidden states \citep{chen2026dflash}, and DFlare changes how drafter layers access those features \citep{zhang2026dflare}. \CPA{} keeps the DFlash backbone and changes two things: how a path is selected from its proposals, and how the backbone is trained.

\paragraph{Correcting parallel drafts.}
\citet{kim2024refinement} show that independent per-position predictions can be improved by searching a top-$k$ lattice with a rescoring language model. DFlash~2 scores adjacent candidate pairs within top-16 shortlists and walks the precomputed scores \citep{inco2026dflash2}; this corresponds to the local term of \CPA{}. Domino corrects the parallel logits with a recurrent head trained under a curriculum \citep{huang2026domino}, and DSpark combines a sequential correction head with confidence-based verification scheduling \citep{cheng2026dspark}. D-Loop reruns the backbone conditioned on a selected prefix \citep{chen2026dloop}, and DDTree verifies a tree of branches rather than a single path \citep{ringel2026ddtree}. \CPA{} sits between local rescoring and recurrent correction. It attends over all earlier choices, as a recurrent head can, but each attention interaction involves one pair of fixed candidates, so every interaction is computed before selection begins and the sequential walk only gathers results.

\paragraph{Training with clean prefixes.}
Domino feeds reference tokens to its correction head during training \citep{huang2026domino}. Joint autoregressive and diffusion objectives have been used to train a model that drafts and verifies its own output \citep{fu2026nemotron}. \CPA{} applies a clean-prefix objective to the shared backbone of a separate drafter for a frozen target. The causal pass is a training signal only and does not run at inference.

\paragraph{Measuring speed.}
Acceptance length and wall-clock speed can disagree: a slower drafter can accept more tokens and still reduce throughput \citep{yan2025decoding}. Following CORAL and DFlash \citep{weng2025coral,chen2026dflash}, we report both quantities at temperatures zero and one, and we relate them through the time a verification round takes.

%% file: sections/03_method.tex
\section{Corrective Prefix Attention}
\label{sec:method}
\CPA{} adds a selector to a parallel block drafter and an auxiliary loss to its training (Figure~\ref{fig:pipeline}). We first describe the proposals the selector chooses from, then the selector, its exact compilation into tables, and finally training and verification.

\subsection{Parallel Proposals and Shortlists}
Let $x$ be the verified context, ending in an anchor token $a$. The drafter proposes $m$ new tokens in a block of $m{+}1$ positions that includes the anchor; we use $m=15$. A trainable backbone $f_\theta$ maps target features, the anchor embedding and $m$ mask embeddings to states $h_{1:m}$ in one parallel pass, and the frozen vocabulary head turns them into unary logits $u_i\in\R^{|\mathcal V|}$. Each position keeps its top-$K$ candidates,
\begin{equation}
 \mathcal C_i=(c_{i,1},\ldots,c_{i,K})=\TopK(u_i),
 \label{eq:shortlist}
\end{equation}
with $K=16$ and the unary argmax in slot~1; we write $\mathcal C_0=(a)$. DFlash proposes $c_{i,1}$ at every position. \CPA{} instead chooses one candidate per position, conditioning on the choices already made in the block. Shortlists are fixed once computed, and no reference or target token is ever inserted into them.

\subsection{Scoring Candidates against the Chosen Prefix}
Let $\hat y_0=a$, and let $\hat y_{1:i-1}$ be the candidates chosen so far in the block. Candidate $v\in\mathcal C_i$ receives the score
\begin{equation}
 S_i(v)=u_i(v)+L_i(v,\hat y_{i-1})+H_i(v).
 \label{eq:score}
\end{equation}
The \emph{local} term measures compatibility with the immediate predecessor,
\begin{equation}
 L_i(v,u)=\frac{\langle c_i\odot p(u),\,s(v)\rangle}{\sqrt r},
 \label{eq:local}
\end{equation}
where $c_i$ is a projection of the draft state $h_i$ and $p(u),s(v)$ are projections of frozen token embeddings, all of rank $r=256$. This is the adjacent-pair correction of DFlash~2 \citep{inco2026dflash2}.

Two paths that end in the same token receive the same local correction, even when their earlier choices differ. The \emph{history} term separates them by letting each candidate attend to positions $0,\ldots,i-2$:
\begin{align}
 H_i(v)&=\sum_{h=1}^{H}\omega_h
 \frac{\sum_{j=0}^{i-2} a^h_{ij}(v,\hat y_j)\,d^h_{ij}(v,\hat y_j)}
 {1+\sum_{j=0}^{i-2}a^h_{ij}(v,\hat y_j)},\label{eq:history}\\
 a^h_{ij}(v,u)&=\exp\!\left(C\tanh\!\left(e^h_{ij}(v,u)/C\right)\right).
 \label{eq:energy}
\end{align}
Here $e^h_{ij}$ is a query--key energy between candidate $v$ at position $i$ and token $u$ at position $j$, with a relative-distance bias; $d^h_{ij}$ is a scalar value read out for that pair; and $\omega_h$ mixes $H=4$ heads of width $r/H=64$. The constant~1 in the denominator acts as a null slot, so a head can attend to nothing, and the bound $C=8$ keeps the weights finite. Appendix~\ref{app:spec} gives the projections. In the example of the introduction, the chosen subject enters the score of \emph{run} through $H_i$, even though the predecessor \emph{often} is the same on both paths.

\subsection{Exact Compilation into Pair Tables}
\label{sec:compile}
The score in Eq.~\ref{eq:history} depends on the chosen prefix, but each pair function depends only on block features and on one pair of (position, token) entries. Because every token that can be chosen is in a shortlist, every pair value can be computed before selection starts. We precompute local tables $T_i[k,\ell]=L_i(c_{i,k},c_{i-1,\ell})$ and, for $0\le j\le i-2$, history tables
\begin{align}
 A^h_{ij}[k,\ell]&=a^h_{ij}(c_{i,k},c_{j,\ell}),\nonumber\\
 B^h_{ij}[k,\ell]&=A^h_{ij}[k,\ell]\,d^h_{ij}(c_{i,k},c_{j,\ell}).
 \label{eq:tables}
\end{align}
Selection keeps a numerator $N_{i,k,h}$, initially $0$, and a denominator $D_{i,k,h}$, initially $1$, for every future position, candidate and head. When position $j$ selects slot $z_j$ (with $z_0=1$ for the anchor), every position $i\ge j+2$ is updated:
\begin{equation}
 N_{i,k,h}\mathrel{+}=B^h_{ij}[k,z_j],\qquad
 D_{i,k,h}\mathrel{+}=A^h_{ij}[k,z_j].
 \label{eq:update}
\end{equation}
Position $i$ then chooses
\begin{equation}
 \begin{split}
 z_i=\argmax_{1\le k\le K}\Bigl[&u_i(c_{i,k})+T_i[k,z_{i-1}]\\
 &+\sum_h\omega_h\frac{N_{i,k,h}}{D_{i,k,h}}\Bigr],
 \end{split}
 \label{eq:choice}
\end{equation}
with ties broken toward the earliest slot. By induction, $N$ and $D$ hold exactly the numerator and denominator of Eq.~\ref{eq:history} before each choice, so compiled and direct selection agree in exact arithmetic (Proposition~1 in Appendix~\ref{app:theory}). Algorithm~\ref{alg:walk} summarizes the procedure.

\begin{algorithm}[t]
\hrule\smallskip
\textbf{Compiled greedy selection}
\begin{enumerate}\setlength\itemsep{1pt}
\item Compute $h$, $u$ and $\mathcal C$, the local tables $T$ and the history tables $A,B$ once for the block, in parallel.
\item Set $z_0=1$, $N=0$, $D=1$, and apply Eq.~\ref{eq:update} for the anchor ($j=0$) to all $i\geq2$.
\item For $i=1,\ldots,m$: choose $z_i$ by Eq.~\ref{eq:choice}; set $\hat y_i=c_{i,z_i}$; apply Eq.~\ref{eq:update} with $j=i$ to all $q\geq i+2$.
\item Send $\hat y_{1:m}$ to the target for verification.
\end{enumerate}
\hrule
\caption{Compiled selection. Only step~3 is sequential; its updates across future positions, candidates and heads run in parallel.}
\label{alg:walk}
\end{algorithm}

\paragraph{Cost.}
Table construction costs $O(m^2K^2Hr_h)$, with $r_h=r/H$, once per block. The walk makes $m$ sequential choices using $K$-way argmax, gathers, additions and lightweight scoring, for $O(m^2KH)$ total work. Construction is quadratic in $K$; the walk is linear. At $m=15$, $K=16$ and $H=4$, packed active history tables contain $188{,}160$ values, under 0.8\,MB in FP32; the implementation instead allocates dense masked tables. The one-layer selector itself is small: with input width $d$ and proposal capacity $m_{\max}$, it adds $5dr+4r^2+H(m_{\max}+2)$ parameters, or 3.5M for Qwen3-4B and 5.5M for Qwen3-8B at $m_{\max}=15$, under $0.1\%$ of either target (Appendix~\ref{app:spec}). Despite this small parameter count, quadratic table construction can make larger shortlists slower (Section~\ref{sec:results}).

\subsection{Training}
\label{sec:training}
\CPA{} trains a drafter with three losses,
\begin{equation}
 \mathcal L=\mathcal L_{\mathrm{DFlash}}+\alpha(t)\,\mathcal L_{\mathrm{sel}}+\lambda_{\mathrm{AR}}\,\mathcal L_{\mathrm{AR}},
 \label{eq:joint}
\end{equation}
where $\mathcal L_{\mathrm{DFlash}}$ is the original masked-block loss over the full vocabulary \citep{chen2026dflash}.

\paragraph{Clean-prefix supervision of the backbone.}
For an anchor $b$ with token $r_{b,0}$ and recorded continuation $r_{b,1:m}$, we run the same backbone causally on the clean tokens, with access to the cached target features $F_{<b}$ that precede the anchor:
\begin{align}
 g_{b,j}&= f_\theta^{\mathrm{causal}}(r_{b,0:j-1}; F_{<b}),\nonumber\\
 \mathcal L_{\mathrm{AR}}
 &=-\frac{\sum_{b,j} V_{b,j}\log \softmax(W_{\mathrm{out}}g_{b,j})_{r_{b,j}}}
 {\max(1,\sum_{b,j} V_{b,j})}.
 \label{eq:ar}
\end{align}
$W_{\mathrm{out}}$ is the frozen vocabulary head, and $V_{b,j}$ excludes padding, document boundaries and positions after the first invalid label. Causal masks in every layer prevent a position from reading its own label. The loss updates the shared backbone but not the selector. It supervises the backbone over the full vocabulary in the condition that verification rewards, predicting the next token from a correct prefix, including tokens the parallel pass leaves out of its shortlist. The clean activations never enter the masked pass, so inference is unchanged. We use $\lambda_{\mathrm{AR}}=0.2$.

\paragraph{Selector training.}
The selector learns to rank each shortlist given a prefix. For a recorded continuation, let $y_i^*=\argmax_v p_T(v\mid x,r_{<i})$ be the cached greedy target token and $S_i^{\mathrm{ref}}$ the score of Eq.~\ref{eq:score} computed with the recorded prefix $r_{<i}$. Where $y_i^*\in\mathcal C_i$, we apply shortlist cross-entropy,
\begin{equation}
 \ell_i=-S_i^{\mathrm{ref}}(y_i^*)+\log\sum_{v\in\mathcal C_i}\exp S_i^{\mathrm{ref}}(v),
 \label{eq:selloss}
\end{equation}
and normalize by the number of valid positions, including those whose target is missing from the shortlist. Once the recorded prefix leaves the greedy target path or the shortlists, its later labels no longer describe what verification would accept, so we gradually down-weight those positions to $\rho=0.25$. The selector-loss weight $\alpha(t)$ is zero through step 2{,}000, then ramps linearly to $0.5$ at step 4{,}000. Both corrections are initialized to zero, so training starts from the unary choices. A replay check rejects selector updates that would shorten correctly selected prefixes on stored training examples. Appendix~\ref{app:guard} gives the schedules and replay details.

\subsection{Verification}
\label{sec:verification}
The target scores the proposed path in one causal forward pass. Under greedy decoding we commit the longest prefix that matches the target's choices, plus the target's token at the first mismatch, or a bonus token if all $m$ proposals match. This reproduces greedy target decoding exactly, whatever the draft quality (Proposition~4). At temperature one, the proposal is still a deterministic path: \CPA{} returns the selected token IDs without draft probabilities. We verify it with the standard rejection sampler of vLLM, which treats each proposal as a point mass. Proposal $\hat y_i$ is accepted with its probability under the processed target distribution; on rejection, $\hat y_i$ is removed from that distribution and a token is sampled from the renormalized remainder. Approximate entropy-based verification and block verification are disabled. This preserves the target distribution \citep{leviathan2023speculative,chen2023speculative} (Appendix~\ref{app:sampling}).

%% file: sections/04_experiments.tex
\section{Experimental Setup}
\label{sec:experiments}
\paragraph{Targets, benchmarks and baselines.}
We use Qwen3-4B and Qwen3-8B \citep{yang2025qwen3} as frozen targets with thinking mode disabled, and evaluate on eight benchmarks: GSM8K, MATH-500 and AIME~2025 for mathematics; HumanEval, MBPP and LiveCodeBench for code; and MT-Bench and Alpaca for dialogue \citep{cobbe2021gsm8k,huggingfaceh42024math500,dekoninck2026matharena,chen2021humaneval,austin2021mbpp,jain2025livecodebench,zheng2023mtbench,taori2023alpaca}. All methods use the same evaluation prompts and decoding settings; Appendix~\ref{app:data} lists the shared dataset versions and splits. We decode at temperatures 0 and 1 with top-$p=1$ and no top-$k$ truncation, generating at most 2{,}048 new tokens per request. We compare with target-only autoregressive decoding, the parallel drafters DFlash \citep{chen2026dflash} and DSpark \citep{cheng2026dspark}, and the autoregressive drafter EAGLE-3 \citep{li2025eagle3}. Every drafter proposes 15 tokens per round, matching block size 16. DSpark's confidence-based scheduling is disabled, so all methods use the same fixed proposal budget.

\paragraph{Training.}
We train all drafters, including EAGLE-3, from scratch for five epochs for each target on the same OpenPerfectBlend prompts with target-generated responses. The \CPA{} drafter uses the DFlash backbone architecture with the target, token embeddings and vocabulary head frozen. The reference configuration uses $K=16$ candidates, rank $r=256$, four heads, one selector layer, $\alpha_{\max}=0.5$ and $\lambda_{\rm AR}=0.2$; Appendix~\ref{app:config} lists all settings. Training uses eight Ascend 910B2 NPUs: six data-parallel training workers and two that generate target hidden states. On Qwen3-4B, one epoch takes 17 hours for DFlash, 23 for DSpark, 28 for \CPA{} and 34 for EAGLE-3. For the selector ablations, we switch off the history term (local only) or the whole selector (parallel argmax) at inference on the \CPA{} checkpoint. For the AR-loss ablation, a drafter with the same architecture and selector is trained for five epochs with $\lambda_{\rm AR}=0$.

\paragraph{Hardware and metrics.}
Decoding runs on one Huawei Ascend 910B2 NPU (64\,GB HBM), using vLLM-Ascend \citep{kwon2023vllm,vllmascend2025}, CANN 9.1.0 and BF16 weights, with one request at a time. Appendix~\ref{app:config} lists software versions. Target verification uses a captured graph; draft backbones run eagerly, and DSpark and \CPA{} correction heads use separate graphs. Speedup $S$ divides output throughput by a fixed AR reference: 47.83 tokens/s for Qwen3-4B and 39.63 tokens/s for Qwen3-8B, at both temperatures. Committed length $\tau$ includes the target correction or bonus token and excludes the anchor; EAGLE-3 pools tokens and rounds within each workload. Workload averages are unweighted, so mean speedup also equals mean throughput divided by the model-specific AR reference. We retain available input precision during aggregation and round final summaries half up to two decimals (Appendix~\ref{app:aggregation}). Absolute speedups depend on the accelerator and kernels \citep{huang2026domino}.

\paragraph{Timing methodology.}
We measure wall-clock stage times using custom instrumentation in the vLLM-Ascend execution path, synchronizing the NPU at each measured stage boundary. Within each workload, we average verification-round timings across all evaluation samples, excluding warmup, prefill and unused final drafts. Table~\ref{tab:latency} reports the unweighted arithmetic mean of the eight workload-level means, without rescaling. ``Other'' is the residual round time outside drafting, correction and verification. Throughput and speedup are measured separately with timing instrumentation disabled.

\input{tables/main_results}

%% file: tables/main_results.tex
\begin{table*}[!t]
\centering
\caption{\textbf{Single-request decoding on Qwen3-4B and Qwen3-8B.} Speedup $S$ uses a fixed AR reference of $47.83$ tokens/s for 4B and $39.63$ tokens/s for 8B at both temperatures. $\tau$ includes the target correction or bonus token; EAGLE-3 uses pooled $\tau$. Avg.\ is an unweighted workload mean. We use full-precision EAGLE-3 logs and the supplied two-decimal workload values for other methods, retain precision during aggregation, and round once (half up). \CPA{} includes prefix attention and AR supervision. LCB: LiveCodeBench.}
\label{tab:main}
\begingroup
\fontsize{8}{9.4}\selectfont
\setlength{\tabcolsep}{1.55pt}
\renewcommand{\arraystretch}{1.13}
\begin{tabular*}{\textwidth}{@{\extracolsep{\fill}}l*{18}{c}@{}}
\toprule
& \multicolumn{6}{c}{\textsc{Math}}
& \multicolumn{6}{c}{\textsc{Code}}
& \multicolumn{4}{c}{\textsc{Chat}}
& \multicolumn{2}{c}{\textsc{Overall}}\\
\cmidrule(lr){2-7}
\cmidrule(lr){8-13}
\cmidrule(lr){14-17}
\cmidrule(lr){18-19}
Method
& \multicolumn{2}{c}{GSM8K}
& \multicolumn{2}{c}{MATH-500}
& \multicolumn{2}{c}{AIME25}
& \multicolumn{2}{c}{HumanEval}
& \multicolumn{2}{c}{MBPP}
& \multicolumn{2}{c}{LCB}
& \multicolumn{2}{c}{MT-Bench}
& \multicolumn{2}{c}{Alpaca}
& \multicolumn{2}{c}{Avg.}\\
\cmidrule(lr){2-3}
\cmidrule(lr){4-5}
\cmidrule(lr){6-7}
\cmidrule(lr){8-9}
\cmidrule(lr){10-11}
\cmidrule(lr){12-13}
\cmidrule(lr){14-15}
\cmidrule(lr){16-17}
\cmidrule(lr){18-19}
& $S$ & $\tau$
& $S$ & $\tau$
& $S$ & $\tau$
& $S$ & $\tau$
& $S$ & $\tau$
& $S$ & $\tau$
& $S$ & $\tau$
& $S$ & $\tau$
& $S$ & $\tau$\\
\midrule
\multicolumn{19}{c}{
\textbf{Qwen3-4B, temperature $=0$} (greedy)}\\
\midrule
EAGLE-3 % S = EAGLE-3 tokens/s / 47.83; Qwen3-4B AR reference
& $1.31\times$ & 2.52
& $1.32\times$ & 2.42
& $1.30\times$ & 2.37
& $1.30\times$ & 2.40
& $1.21\times$ & 2.27
& $1.24\times$ & 2.30
& $1.17\times$ & 2.17
& $1.18\times$ & 2.20
& $1.25\times$ & 2.33\\
DFlash
& $3.95\times$ & 7.76
& $4.27\times$ & 7.72
& $3.49\times$ & 6.57
& $3.11\times$ & 5.91
& $2.77\times$ & 5.31
& $3.02\times$ & 5.79
& $1.66\times$ & 3.33
& $1.69\times$ & 3.19
& $3.00\times$ & 5.70\\
DSpark
& $4.04\times$ & 8.28
& $4.34\times$ & 8.19
& $3.67\times$ & 7.20
& $3.31\times$ & 6.58
& $3.00\times$ & 6.02
& $3.09\times$ & 6.18
& $1.76\times$ & 3.78
& $1.77\times$ & 3.66
& $3.12\times$ & 6.24\\
\textbf{CPA}
& $4.38\times$ & 9.85
& $4.68\times$ & 9.62
& $3.99\times$ & 8.05
& $3.49\times$ & 7.17
& $3.12\times$ & 6.61
& $3.29\times$ & 6.71
& $1.98\times$ & 5.12
& $1.88\times$ & 3.95
& $3.35\times$ & 7.14\\
\midrule
\multicolumn{19}{c}{
\textbf{Qwen3-4B, temperature $=1$} (sampling)}\\
\midrule
EAGLE-3 % S = EAGLE-3 tokens/s / 47.83; Qwen3-4B AR reference
& $1.27\times$ & 2.46
& $1.19\times$ & 2.23
& $1.10\times$ & 2.07
& $1.24\times$ & 2.35
& $1.16\times$ & 2.21
& $1.08\times$ & 2.04
& $1.09\times$ & 2.06
& $1.12\times$ & 2.12
& $1.16\times$ & 2.19\\
DFlash
& $3.54\times$ & 7.02
& $3.67\times$ & 6.84
& $2.41\times$ & 4.85
& $2.79\times$ & 5.35
& $2.46\times$ & 4.81
& $2.88\times$ & 5.47
& $1.45\times$ & 3.02
& $1.52\times$ & 2.92
& $2.59\times$ & 5.04\\
DSpark
& $3.68\times$ & 7.65
& $3.76\times$ & 7.28
& $2.74\times$ & 5.61
& $2.96\times$ & 6.07
& $2.66\times$ & 5.42
& $2.85\times$ & 5.76
& $1.57\times$ & 3.46
& $1.60\times$ & 3.38
& $2.73\times$ & 5.58\\
\textbf{CPA}
& $3.79\times$ & 8.74
& $4.04\times$ & 8.60
& $2.91\times$ & 6.22
& $3.14\times$ & 6.58
& $2.60\times$ & 5.73
& $3.07\times$ & 6.36
& $1.65\times$ & 4.41
& $1.70\times$ & 3.68
& $2.86\times$ & 6.29\\
\midrule
\multicolumn{19}{c}{
\textbf{Qwen3-8B, temperature $=0$} (greedy)}\\
\midrule
EAGLE-3 % S = EAGLE-3 tokens/s / 39.63; Qwen3-8B AR reference
& $1.51\times$ & 2.47
& $1.47\times$ & 2.37
& $1.41\times$ & 2.28
& $1.52\times$ & 2.48
& $1.33\times$ & 2.18
& $1.32\times$ & 2.14
& $1.26\times$ & 2.05
& $1.19\times$ & 1.96
& $1.38\times$ & 2.24\\
DFlash
& $4.25\times$ & 7.93
& $4.28\times$ & 7.59
& $3.31\times$ & 6.38
& $3.37\times$ & 6.14
& $2.69\times$ & 5.24
& $3.24\times$ & 6.02
& $1.79\times$ & 3.41
& $1.60\times$ & 3.09
& $3.07\times$ & 5.73\\
DSpark
& $4.09\times$ & 8.14
& $4.68\times$ & 8.47
& $3.57\times$ & 7.03
& $3.45\times$ & 6.72
& $3.05\times$ & 6.27
& $3.06\times$ & 6.03
& $1.76\times$ & 3.66
& $1.85\times$ & 3.87
& $3.19\times$ & 6.27\\
\textbf{CPA}
& $4.27\times$ & 9.51
& $5.03\times$ & 9.97
& $3.74\times$ & 7.82
& $3.72\times$ & 7.48
& $2.84\times$ & 6.35
& $3.48\times$ & 7.03
& $2.01\times$ & 5.27
& $1.89\times$ & 4.08
& $3.37\times$ & 7.19\\
\midrule
\multicolumn{19}{c}{
\textbf{Qwen3-8B, temperature $=1$} (sampling)}\\
\midrule
EAGLE-3 % S = EAGLE-3 tokens/s / 39.63; Qwen3-8B AR reference
& $1.37\times$ & 2.38
& $1.28\times$ & 2.19
& $1.17\times$ & 2.00
& $1.34\times$ & 2.31
& $1.21\times$ & 2.10
& $1.10\times$ & 1.90
& $1.11\times$ & 1.91
& $1.12\times$ & 1.90
& $1.21\times$ & 2.09\\
DFlash
& $3.86\times$ & 7.31
& $3.61\times$ & 6.55
& $2.24\times$ & 4.66
& $2.86\times$ & 5.42
& $2.29\times$ & 4.61
& $2.97\times$ & 5.63
& $1.55\times$ & 3.08
& $1.45\times$ & 2.87
& $2.60\times$ & 5.02\\
DSpark
& $3.69\times$ & 7.48
& $4.07\times$ & 7.59
& $2.54\times$ & 5.29
& $3.09\times$ & 6.18
& $2.60\times$ & 5.53
& $2.86\times$ & 5.69
& $1.56\times$ & 3.39
& $1.60\times$ & 3.43
& $2.75\times$ & 5.57\\
\textbf{CPA}
& $3.69\times$ & 8.47
& $4.34\times$ & 8.83
& $2.69\times$ & 6.01
& $3.26\times$ & 6.79
& $2.42\times$ & 5.66
& $3.14\times$ & 6.46
& $1.66\times$ & 4.57
& $1.67\times$ & 3.73
& $2.86\times$ & 6.32\\
\bottomrule
\end{tabular*}
\endgroup
\end{table*}

%% file: sections/05_results.tex
\section{Results and Analysis}
\label{sec:results}
\input{tables/sensitivity_training}
\paragraph{CPA is fastest on average in every setting.}
Table~\ref{tab:main} reports all four combinations of target size and temperature. \CPA{} has the highest mean speedup and the longest mean committed length in each: $3.35\times$ and $3.37\times$ with greedy decoding on Qwen3-4B and Qwen3-8B, and $2.86\times$ at temperature one on both. Using unrounded computed workload means, its speedup exceeds DSpark's by 3.9--7.3\% and DFlash's by 9.8--11.9\%, and it commits 12.7--14.6\% more tokens per round than DSpark and about 25\% more than DFlash. The gains are similar for the two target sizes. EAGLE-3, which drafts one token per step, commits 2.1--2.3 tokens per round, with mean speedups of $1.25\times$, $1.16\times$, $1.38\times$ and $1.21\times$ in table order.

Per benchmark, \CPA{} commits the most tokens in all 32 cases and is fastest in 28. The exceptions are MBPP in three settings, where its gain in $\tau$ over DSpark falls to 1--6\%, and GSM8K on Qwen3-8B at temperature one, where DFlash is fastest despite committing 14\% fewer tokens. The largest gains in $\tau$ appear on MT-Bench (27--44\% over DSpark), but there the speed gain (5--14\%) is in line with the other benchmarks.

\input{tables/latency}
\paragraph{Longer commits cost more per round.}
For a fixed AR reference $t_{\mathrm{AR}}$, $S\approx\tau\,t_{\mathrm{AR}}/t_{\mathrm{round}}$, so $\tau/S$ is an inferred round-cost index in units of the reference token time. Averaging this index over the 32 workload settings gives 1.92 for DFlash, 2.03 for DSpark and 2.20 for \CPA{}. The \CPA{} index is about 15\% higher than DFlash's and 8\% higher than DSpark's, alongside gains of about 25\% and 14\% in committed length. The indices reflect all overheads included in the throughput measurements and explain why acceptance gains need not translate proportionally into speed; they are not independently measured round latencies. Table~\ref{tab:latency} reports synchronized wall-clock stage timings from the instrumented runs. \CPA{} spends 2.748\,ms on correction, 6.18\% of its 44.499\,ms mean instrumented round time; drafting and target verification take 14.138 and 13.406\,ms, respectively.

\input{figures/tradeoff}
\paragraph{More acceptance does not always mean more speed.}
Table~\ref{tab:ablation} varies the loss weights and the selector's capacity around the reference configuration, and Figure~\ref{fig:tradeoff} plots each variant's committed length against its speed. A shortlist of 32 raises $\tau$ from $7.14$ to $7.19$ but lowers speedup from $3.35\times$ to $3.26\times$, consistent with table construction growing quadratically in $K$. Rank 512 similarly raises $\tau$ to $7.31$ but gives only $3.26\times$, and the stacked two-layer variant gives $\tau=7.22$ and $3.34\times$ (Appendix~\ref{app:stacked}). Smaller settings lose on both metrics: $K=8$ and $r=128$ reach only $3.19\times$. These higher-acceptance variants sit on a higher-cost-index line in Figure~\ref{fig:tradeoff}: its extra acceptance is paid for in time per round. The reference configuration ($K=16$, $r=256$, one layer) is therefore the fastest of those tested, not the most accurate. Raising the selector-loss weight $\alpha_{\max}$ from $0.125$ to $0.5$ increases $\tau$ from $6.58$ to $7.14$ and speedup from $3.09\times$ to $3.35\times$, while $\alpha_{\max}=1$ lowers both.

\input{tables/component_ablation}
\paragraph{Where the gain comes from.}
Table~\ref{tab:components} decomposes the gain on Qwen3-4B with greedy decoding. Even with its selector switched off, the \CPA{} checkpoint commits more tokens than DFlash (6.26 against 5.70), so part of the gain comes from training rather than selection. The local term raises $\tau$ by a further 8.4\% and speed by 2.1\%; its inferred cost index rises from $2.00$ to $2.12$. Adding history raises $\tau$ by 5.1\% and speed by 4.8\%, with a similar inferred cost index ($2.13$). These indices are ratios of workload means, not independently measured round latencies. History improves both metrics on all eight benchmarks, most on MT-Bench (14.8\% in $\tau$) and least on LiveCodeBench and Alpaca (2.0--2.3\%; Table~\ref{tab:extended}). Dependencies beyond the previous token thus account for most of the selector's speed gain.

Adding the AR loss raises mean speedup from $3.28\times$ to $3.35\times$ and mean $\tau$ from $6.93$ to $7.14$: gains of 2.1\% and 3.0\%, respectively, using the unrounded computed means. Both metrics improve on every benchmark. The inference path is identical, so this gain costs nothing at decoding time. Speedup is flat for AR weights between $0.1$ and $0.4$ ($3.34$--$3.35\times$; Table~\ref{tab:ablation}), so the method is not sensitive to the exact value of $\lambda_{\rm AR}$.

\input{tables/block_size}
\paragraph{Block size and concurrency.}
Block size 16 gives the highest $\tau$ and speedup for all three parallel drafters (Table~\ref{tab:block-size}). Longer blocks offer more positions to accept, but late positions rarely survive while drafting and table construction cost more; at $B=32$ every method commits fewer tokens and runs more slowly than at $B=16$. The gain of \CPA{} over DSpark in $\tau$ is smallest at $B=8$ (4.5\%), where there are fewer earlier choices to attend to, and larger at $B=16$ and $32$ (14.4\% and 10.3\%). Under concurrent requests, speedup over target-only decoding shrinks as batching makes the target more efficient: \CPA{} goes from $3.35\times$ with one request to $1.52\times$ with 32, compared with $1.42\times$ for DSpark at 32 (Table~\ref{tab:serving-sensitivity} in Appendix~\ref{app:sensitivity}).

%% file: tables/sensitivity_training.tex
\begin{table*}[!t]
\centering
\caption{\textbf{Loss weights and selector capacity.} Greedy Qwen3-4B; $S$ and $\tau$ are eight-benchmark means. Bold rows are the reference configuration ($\alpha_{\max}=0.5$, $\lambda_{\rm AR}=0.2$, $r=256$, $K=16$, one selector layer), recomputed from the Table~\ref{tab:main} workload values and repeated in each panel. Other aggregate-only observations retain their supplied precision. Larger $K$ or $r$ raises $\tau$ but not speed. The stacked two-layer row uses \texttt{local\_prefix\_v2} (Appendix~\ref{app:stacked}).}
\label{tab:training-sensitivity}\label{tab:ablation}
\begingroup
\small
\setlength{\tabcolsep}{4pt}
\renewcommand{\arraystretch}{1.12}
\begin{minipage}[t]{.47\textwidth}
\vspace{0pt}
\centering
\begin{tabular*}{\linewidth}{@{\extracolsep{\fill}}lrr@{}}
\toprule
Setting & Speedup $S$ & Tokens/round $\tau$\\
\midrule
\multicolumn{3}{@{}l}{\textit{Selector-loss weight, $\alpha_{\max}$}}\\
$0.125$ & $3.09\times$ & $6.58$\\
$0.250$ & $3.20\times$ & $6.88$\\
$\mathbf{0.500}$ & $\mathbf{3.35}\times$ & $\mathbf{7.14}$\\
$1.000$ & $3.30\times$ & $7.03$\\
\midrule
\multicolumn{3}{@{}l}{\textit{AR-loss weight, $\lambda_{\rm AR}$}}\\
$0$ (without AR loss) & $3.28\times$ & $6.93$\\
$0.1$ & $3.34\times$ & $7.09$\\
$\mathbf{0.2}$ & $\mathbf{3.35}\times$ & $\mathbf{7.14}$\\
$0.4$ & $3.35\times$ & $7.11$\\
\midrule
\multicolumn{3}{@{}l}{\textit{Selector implementation}}\\
\textbf{Compiled (1 layer)} & $\mathbf{3.35}\times$ & $\mathbf{7.14}$\\
Stacked v2 (2 layers) & $3.34\times$ & $7.22$\\
\bottomrule
\end{tabular*}
\end{minipage}\hfill
\begin{minipage}[t]{.47\textwidth}
\vspace{0pt}
\centering
\begin{tabular*}{\linewidth}{@{\extracolsep{\fill}}lrr@{}}
\toprule
Setting & Speedup $S$ & Tokens/round $\tau$\\
\midrule
\multicolumn{3}{@{}l}{\textit{Shortlist size, $K$}}\\
$8$ & $3.19\times$ & $6.98$\\
$\mathbf{16}$ & $\mathbf{3.35}\times$ & $\mathbf{7.14}$\\
$32$ & $3.26\times$ & $7.19$\\
\midrule
\multicolumn{3}{@{}l}{\textit{Total selector rank, $r$}}\\
$128$ & $3.19\times$ & $6.88$\\
$\mathbf{256}$ & $\mathbf{3.35}\times$ & $\mathbf{7.14}$\\
$512$ & $3.26\times$ & $7.31$\\
\bottomrule
\end{tabular*}
\smallskip
\parbox{\linewidth}{\footnotesize
$\alpha_{\max}$ is the final weight of the scheduled selector loss; it is distinct from $\lambda_{\rm AR}$. Rank denotes the total projection width, split across four heads. $K$ is the number of candidates per position, not the number of proposed positions.}
\end{minipage}
\endgroup
\end{table*}

%% file: tables/latency.tex
\begin{table}[t]
\centering
\caption{\textbf{Instrumented per-round timing breakdown.} Greedy Qwen3-4B, one request; milliseconds. Custom wall-clock instrumentation synchronizes the NPU at stage boundaries. We average verification-round times across all evaluation samples within each workload, then average the eight workload means equally, without rescaling. Warmup, prefill and unused final drafts are excluded. Other is the residual round time. Throughput is measured separately with instrumentation disabled. The accompanying $\tau$ values retain their reported precision.}
\label{tab:latency}
\begingroup
\fontsize{8.5}{10}\selectfont
\setlength{\tabcolsep}{2.5pt}
\renewcommand{\arraystretch}{1.12}
\begin{tabular*}{\columnwidth}{@{\extracolsep{\fill}}lrrrrrr@{}}
\toprule
Method & Draft & Correction & Verify & Other & Total & $\tau$\\
\midrule
DFlash & 13.480 & -- & 13.158 & 13.084 & 39.7222 & 5.70\\
DSpark & 13.752 & 1.177 & 13.203 & 13.682 & 41.815 & 6.24\\
\CPA{} & 14.138 & 2.748 & 13.406 & 14.207 & 44.499 & \textbf{7.13}\\
\bottomrule
\end{tabular*}
\endgroup
\end{table}

%% file: figures/tradeoff.tex
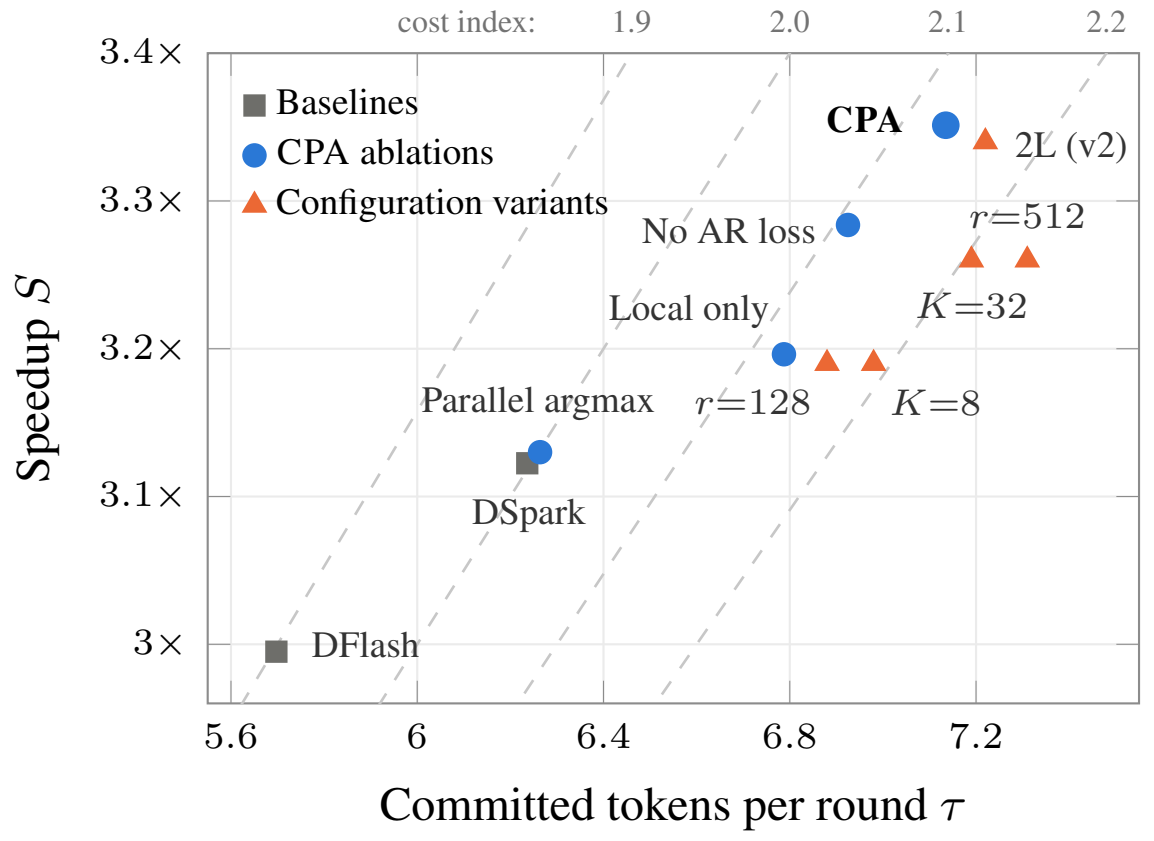
\begin{figure}[t]
\centering
\begin{tikzpicture}
\definecolor{cpaBlue}{HTML}{2A78D6}
\definecolor{capOrange}{HTML}{EB6834}
\definecolor{baseGray}{HTML}{6E6E6A}
\begin{axis}[
  scale only axis, width=0.80\columnwidth, height=4.3cm,
  xmin=5.55, xmax=7.55, ymin=2.96, ymax=3.40,
  xlabel={Committed tokens per round $\tau$},
  ylabel={Speedup $S$},
  xlabel style={font=\small}, ylabel style={font=\small},
  tick label style={font=\scriptsize},
  ytick={3.0,3.1,3.2,3.3,3.4},
  yticklabel={$\pgfmathprintnumber[fixed,precision=1]{\tick}\times$},
  xtick={5.6,6.0,6.4,6.8,7.2},
  axis line style={black!45}, tick style={black!45},
  grid=major, grid style={black!8},
  clip mode=individual,
  legend style={font=\scriptsize, draw=none, fill=none, at={(0.02,0.98)}, anchor=north west, row sep=-1pt},
  legend cell align=left,
]
% iso-cost-index guides: S = tau / c
\foreach \c/\lab in {1.9/1.9, 2.0/2.0, 2.1/2.1, 2.2/2.2} {
  \edef\tmp{\noexpand\addplot[black!22, dashed, line width=0.5pt, domain=5.55:7.55, samples=2, forget plot] {x/\c};}
  \tmp
}
\node[font=\tiny, text=black!55, anchor=south] at (axis cs:6.46,3.40) {1.9};
\node[font=\tiny, text=black!55, anchor=south] at (axis cs:6.80,3.40) {2.0};
\node[font=\tiny, text=black!55, anchor=south] at (axis cs:7.14,3.40) {2.1};
\node[font=\tiny, text=black!55, anchor=south] at (axis cs:7.48,3.40) {2.2};
\node[font=\tiny, text=black!55, anchor=south east] at (axis cs:6.30,3.40) {cost index:};

\addplot[only marks, mark=square*, mark size=1.9pt, color=baseGray]
  coordinates {(5.6975,2.995) (6.23625,3.1225)};
\addlegendentry{Baselines}
\addplot[only marks, mark=*, mark size=2.1pt, color=cpaBlue]
  coordinates {(6.26375,3.13) (6.7875,3.19625) (6.925,3.28375)};
\addlegendentry{\CPA{} ablations}
\addplot[only marks, mark=triangle*, mark size=2.4pt, color=capOrange]
  coordinates {(6.98,3.19) (6.88,3.19) (7.19,3.26) (7.31,3.26) (7.22,3.34)};
\addlegendentry{Configuration variants}
\addplot[only marks, mark=*, mark size=3.0pt, color=cpaBlue, draw=white, line width=0.8pt, forget plot]
  coordinates {(7.135,3.35125)};

\begin{scope}[font=\scriptsize, text=black!80]
\node[anchor=west] at (axis cs:5.73,3.00) {DFlash};
\node[anchor=north] at (axis cs:6.24,3.112) {DSpark};
\node[anchor=south] at (axis cs:6.26,3.138) {Parallel argmax};
\node[anchor=south east] at (axis cs:6.80,3.200) {Local only};
\node[anchor=east] at (axis cs:6.90,3.28) {No AR loss};
\node[anchor=north east] at (axis cs:6.89,3.186) {$r{=}128$};
\node[anchor=north west] at (axis cs:6.97,3.186) {$K{=}8$};
\node[anchor=north] at (axis cs:7.19,3.252) {$K{=}32$};
\node[anchor=south] at (axis cs:7.31,3.268) {$r{=}512$};
\node[anchor=west] at (axis cs:7.24,3.336) {2L (v2)};
\node[anchor=east, font=\scriptsize\bfseries, text=black] at (axis cs:7.09,3.355) {\CPA{}};
\end{scope}
\end{axis}
\end{tikzpicture}
\caption{\textbf{Acceptance and speed are not the same.} Greedy Qwen3-4B means from Tables~\ref{tab:main}, \ref{tab:ablation} and~\ref{tab:components}. Dashed lines show equal inferred cost index $\bar\tau/\bar S$. Orange points show configuration variants; the stacked variant is specified in Appendix~\ref{app:stacked}. Coordinates use unrounded computed means where workload values are available.}
\label{fig:tradeoff}
\end{figure}

%% file: tables/component_ablation.tex
\begin{table}[t]
\centering
\caption{\textbf{Where the gain comes from.} Greedy Qwen3-4B, eight-benchmark means derived from Table~\ref{tab:extended}. Parallel argmax and Local only disable the selector or its history term at inference; No AR loss is trained with $\lambda_{\rm AR}=0$. The cost index is the ratio of unrounded workload means $\bar\tau/\bar S$, an inferred normalized cost rather than a measured mean round latency.}
\label{tab:components}
\begingroup
\small
\setlength{\tabcolsep}{4pt}
\renewcommand{\arraystretch}{1.12}
\begin{tabular*}{\columnwidth}{@{\extracolsep{\fill}}lrrr@{}}
\toprule
Variant & $S$ & $\tau$ & Index\\
\midrule
DFlash (reference) & $3.00\times$ & $5.70$ & $1.90$\\
\midrule
\CPA{}, parallel argmax & $3.13\times$ & $6.26$ & $2.00$\\
\CPA{}, local term only & $3.20\times$ & $6.79$ & $2.12$\\
\textbf{\CPA{}, local + history} & $\mathbf{3.35}\times$ & $\mathbf{7.14}$ & $2.13$\\
\midrule
\CPA{} without AR loss & $3.28\times$ & $6.93$ & $2.11$\\
\bottomrule
\end{tabular*}
\endgroup
\end{table}

%% file: tables/block_size.tex
\begin{table}[t]
\centering
\caption{\textbf{Block size changes both acceptance and cost.} Greedy Qwen3-4B. Each method reports committed tokens per round $\tau$ and speedup $S$. $B$ includes the anchor.}
\label{tab:block-size}
\begingroup
\fontsize{8.4}{10}\selectfont
\setlength{\tabcolsep}{2.2pt}
\renewcommand{\arraystretch}{1.12}
\begin{tabular*}{\columnwidth}{@{\extracolsep{\fill}}lrrrrrr@{}}
\toprule
& \multicolumn{2}{c}{DFlash} & \multicolumn{2}{c}{DSpark} & \multicolumn{2}{c}{CPA}\\
\cmidrule(lr){2-3}\cmidrule(lr){4-5}\cmidrule(lr){6-7}
$B$ & $\tau$ & $S$ & $\tau$ & $S$ & $\tau$ & $S$\\
\midrule
$8$ & $4.332$ & $2.37\times$ & $5.019$ & $2.58\times$ & $5.247$ & $2.61\times$\\
$16$ & $5.698$ & $3.00\times$ & $6.236$ & $3.12\times$ & $7.135$ & $3.35\times$\\
$32$ & $4.871$ & $1.69\times$ & $5.605$ & $1.85\times$ & $6.182$ & $1.92\times$\\
\bottomrule
\end{tabular*}
\smallskip
\parbox{\columnwidth}{\footnotesize
$B=8,16,32$ permit at most $7,15,31$ proposed tokens; the anchor is excluded from $\tau$. Each block size uses separately trained drafters. The $B=16$ row is recomputed from Table~\ref{tab:main}; other rows retain the supplied aggregate values.}
\endgroup
\end{table}

%% file: sections/06_conclusion.tex
\section{Conclusion}
Parallel drafters are fast because they score every position at once, but verification rewards a block that is consistent from its first token onward. \CPA{} reconciles the two. Its selector attends over every earlier choice in the block, and because the shortlists are fixed, all of that attention can be computed before selection begins, leaving a sequential walk of lookups and additions. An auxiliary autoregressive loss gives the shared backbone practice at predicting from correct prefixes at no inference cost. Attention to older choices accounts for most of the selector's speed gain while keeping inferred round cost similar. Across eight benchmarks, two target sizes and two temperatures, \CPA{} commits 13--15\% more tokens per round than DSpark and has the highest mean speedup in every setting. The capacity sweeps carry a broader lesson for draft correction: added context helps only as long as it stays cheaper than the tokens it saves.

%% file: sections/07_limitations.tex
\section*{Limitations}
\paragraph{Scope of the evaluation.}
We evaluate two targets from one model family, Qwen3-4B and Qwen3-8B, and run the ablations on Qwen3-4B only. The reported throughput results use one accelerator type, the Ascend 910B2, with vLLM-Ascend. The stage breakdown uses synchronization at each measured boundary and characterizes the instrumented execution; throughput and speedup use separate runs with instrumentation disabled. All methods use a fixed model-specific AR throughput reference, so individual workload ratios are relative to that reference rather than to a separately measured AR throughput for that workload. The relative cost of the sequential selection walk depends on kernel-launch overheads and may differ on GPUs or other serving stacks, so the speedups should not be transferred directly to other hardware. We report single training runs and do not give confidence intervals over retraining or repeated timing; sub-percent differences, such as $3.35\times$ against $3.34\times$, do not establish a reliable ranking without repeated measurements. Appendix~\ref{app:config} documents the training recipe and seed roles; final resumption budgets, training-data revisions and evaluated checkpoint identifiers are not available in the accompanying run metadata.

\paragraph{Method.}
The selector can only choose among the top-$K$ candidates of the parallel pass, so a target token outside the shortlist ends the accepted prefix regardless of the selector. Table construction grows quadratically with the shortlist size and the block length, which limits how far either can be increased; our capacity sweeps show this cost overtaking the acceptance gain at $K=32$. The no-AR drafter was trained in a separate run, so factors other than $\lambda_{\rm AR}$ may contribute to the gap. The auxiliary causal pass and the selector make training more expensive: on Qwen3-4B, one \CPA{} epoch takes 28 hours on our hardware, against 17 for DFlash and 23 for DSpark. The AR-loss ablation compares drafters at the same number of epochs, not the same compute budget. The selector is trained on recorded prefixes rather than its own choices, so some exposure mismatch remains.

\paragraph{Exactness.}
The equivalence of compiled and direct selection, and of greedy speculative and target decoding, holds in exact arithmetic. In BF16, changes in summation order can flip decisions near ties, so deployments should check output parity on their own kernels.

\section*{Ethical Considerations}
\CPA{} accelerates a fixed target model and does not change what it generates under greedy decoding or, with the sampler we use, its output distribution. It therefore neither mitigates nor adds to the target's risks, such as factual errors or biased outputs. Faster decoding can lower the energy cost per generated token, but we did not measure training or serving energy. All benchmarks are public and used under their licenses.

%% file: sections/appendix.tex
% Appendix float placement: let tables share pages with text and sit at the top of float pages.
\renewcommand{\topfraction}{0.9}
\renewcommand{\dbltopfraction}{0.9}
\renewcommand{\textfraction}{0.07}
\renewcommand{\floatpagefraction}{0.85}
\renewcommand{\dblfloatpagefraction}{0.85}
\makeatletter
\setlength{\@fptop}{0pt}\setlength{\@fpbot}{0pt plus 1fil}
\setlength{\@dblfptop}{0pt}\setlength{\@dblfpbot}{0pt plus 1fil}
\makeatother

\input{sections/appendix_theory}

\section{Selector Specification}
\label{app:spec}
Let $d$ be the common width of the draft states and frozen token embeddings, $r$ the total projection rank and $r_h=r/H$ the width of each head. For a vector $x$, define $\nu(x)=x/\sqrt{\operatorname{mean}(x^2)+10^{-6}}$, an RMS normalization without learned parameters. The local score in Eq.~\ref{eq:local} uses
\begin{align}
 F(v)&=\nu(E(v)), & c_i&=\nu(W_h\nu(h_i)),\\
 p(u)&=\nu(W_pF(u)), & s(v)&=W_sF(v),
\end{align}
where $E$ is the frozen embedding table. The candidate and memory features for the history term are
\begin{align}
 z_i(v)&=\nu(c_i+W_cF(v)),\\
 m(u)&=\nu(W_mF(u)).
\end{align}
The $h$th slices of $W_qz_i(v)$ and $W_tz_i(v)$ are $q_i^h(v)$ and $t_i^h(v)$; the corresponding slices of $W_km(u)$ and $W_vm(u)$ are $k^h(u)$ and $v^h(u)$. For $j\leq i-2$, the energy and readout are
\begin{align}
 e_{ij}^h(v,u)&=\frac{\langle q_i^h(v),k^h(u)\rangle}{\sqrt{r_h}}+b^h_{i-j},
 \label{eq:proposedenergy}\\
 d_{ij}^h(v,u)&=\frac{\langle t_i^h(v),v^h(u)\rangle}{\sqrt{r_h}}.
 \label{eq:proposedreadout}
\end{align}
The five input maps are $d\!\to\!r$ and the four history maps are $r\!\to\!r$, all without bias. With maximum proposal length $m_{\max}$, the one-layer selector has $5dr+4r^2+H(m_{\max}+2)$ parameters, including the distance biases and the $H$ mixing weights. For the reference configuration $r=256$, $H=4$ and $m_{\max}=m=15$, the exact counts are $3{,}539{,}012$ for Qwen3-4B ($d=2560$) and $5{,}505{,}092$ for Qwen3-8B ($d=4096$). These counts include only the added selector, excluding the draft backbone and frozen embeddings and vocabulary head. We use Xavier-uniform initialization and then set $W_s$, $b$ and $\omega$ to zero, so that initial choices follow the unary logits (Proposition~3). Table~\ref{tab:shapes} lists the per-block tensor shapes.

\begin{table}[H]
\centering\smalltable
\begin{tabular}{ll}
\toprule
Object & Per-block shape\\
\midrule
Unary logits & $m\times|\mathcal V|$\\
Candidate IDs & $m\times K$\\
Candidate embeddings & $m\times K\times d$\\
Local table, $i>1$ & $K\times K$\\
Local table, $i=1$ & $K\times1$\\
History table, $j>0$ & $H\times K\times K$\\
History table, $j=0$ & $H\times K\times1$\\
Each accumulator & $m\times K\times H$\\
Selected proposal IDs & $m$\\
\bottomrule
\end{tabular}
\caption{Per-block tensor shapes. Our implementation stores dense FP32 history tables, masks pairs with $j>i-2$, and handles the anchor separately.}
\label{tab:shapes}
\end{table}

\paragraph{Stacked cross-attention implementation.}
\label{app:stacked}
The \texttt{local\_prefix\_v2} implementation supports a stacked variant with two independently parameterized cross-attention blocks. At rank 256, each block uses pre-attention query LayerNorm and four 64-dimensional heads, with query, key, value and output projections followed by a residual connection. A pre-FFN LayerNorm and a $256\!\to\!512\!\to\!256$ feed-forward network with SiLU form a second residual branch. Let $q^{(0)}$ be the candidate query and $M$ the older selected-prefix memory. The blocks operate sequentially:
\begin{equation}
 q^{(1)}=F_1(q^{(0)},M),\qquad
 q^{(2)}=F_2(q^{(1)},M).
\end{equation}
Both blocks attend to the same memory, which is not updated between layers. The memory excludes the immediate predecessor, whose contribution remains in the separate local term. The history correction is
\begin{equation}
 h=w_{\mathrm{out}}^\top\bigl(q^{(2)}-q^{(0)}\bigr),
\end{equation}
and is added to the unary score and local correction. These architectural details apply to \texttt{local\_prefix\_v2}. The \texttt{compiled\_prefix\_v3} implementation contains one history layer; setting its layer field to two does not instantiate this stack. The second block's query depends on the selected prefix through the first block, so the fixed-pair compilation proof and lookup-walk cost bounds do not apply to this stacked variant. The reported two-layer result ($\tau=7.22$, $S=3.34\times$) uses \texttt{local\_prefix\_v2}. Its comparison with the compiled one-layer reference therefore changes the selector architecture and execution path as well as depth; it does not isolate depth alone.

\section{Training Details}
\label{app:guard}
\label{app:config}
\begin{table*}[t]
\centering\small
\setlength{\tabcolsep}{5pt}
\begin{tabular*}{\textwidth}{@{\extracolsep{\fill}}p{0.38\textwidth}p{0.58\textwidth}@{}}
\toprule
Setting & Value\\
\midrule
\multicolumn{2}{@{}l}{\textit{Reference configuration}}\\
Recipe epoch target & 5; resumption may override the final budget\\
Final selector-loss weight $\alpha_{\max}$ & $0.5$\\
AR-loss weight $\lambda_{\rm AR}$ & $0.2$ ($0$ for the no-AR ablation)\\
Block size / proposed tokens / shortlist size & $16$ / $15$ / $16$\\
Selector rank / heads / layers & $256$ / $4$ / $1$\\
\midrule
\multicolumn{2}{@{}l}{\textit{Schedules and implementation}}\\
Frozen components & Target, token embeddings, vocabulary head\\
Energy bound $C$ / local and history scales & $8$ / $1$ and $1$\\
Selector-loss delay / ramp & $2{,}000$ / $2{,}000$ optimizer steps\\
Prefix-weight start / ramp / floor $\rho$ & $4{,}000$ / $4{,}000$ steps / $0.25$\\
Replay guard start / buffer / additions & Step $4{,}000$ / $32$ / $4$ anchors per worker\\
Guard trial scales & $1,\tfrac12,\tfrac14,\tfrac18$\\
AR anchors per worker / logit chunk & $64$ / $128$ positions\\
Precision & FP32 trainable parameters; BF16 autocast and export\\
Gradient synchronization / clipping & 6 DDP ranks by default / global norm $1.0$\\
Packed positions per rank / distributed step & $3{,}072$ / $18{,}432$ at 6 DDP ranks\\
\midrule
\multicolumn{2}{@{}l}{\textit{Optimizers and learning-rate schedules}}\\
Eligible trainable backbone 2D matrices & Muon; learning rate $6\times10^{-3}$\\
Muon momentum / weight decay & $0.95$ / $0.1$\\
Muon orthogonalization & 5 Newton--Schulz iterations\\
Muon \texttt{adjust\_lr\_fn} & \texttt{match\_rms\_adamw}\\
Remaining trainable backbone parameters & AdamW; learning rate $6\times10^{-4}$; decay $0.01$\\
Backbone learning-rate schedule & Cosine decay; warmup over $2\%$ of planned steps\\
Selector optimizer / peak learning rate & Separate AdamW / $3\times10^{-4}$\\
Selector weight decay & $0.01$ for parameters with $\geq2$ dimensions; $0$ otherwise\\
Selector learning-rate schedule & $2{,}000$-step delay, $2{,}000$-step linear warmup, cosine decay\\
\midrule
\multicolumn{2}{@{}l}{\textit{Hardware and software}}\\
Training hardware & 8 Ascend 910B2 (6 training, 2 target-feature)\\
Time per epoch, Qwen3-4B & EAGLE-3 34\,h; DFlash 17\,h; DSpark 23\,h; \CPA{} 28\,h\\
vLLM & \texttt{0.1.dev20548+g77b1d88c5}\\
vLLM-Ascend & \texttt{0.1.dev5188+g276a5c840}\\
Speculators / PyTorch / torch-npu & \texttt{0.8.0.dev0} / \texttt{2.10.0} / \texttt{2.10.0.post4}\\
CANN & \texttt{9.1.0}\\
\bottomrule
\end{tabular*}
\caption{Reference configuration and training recipe. The selector-loss weight $\alpha_{\max}$, prefix-weight floor $\rho$, and optimizer learning-rate schedules are distinct. Recipe defaults and remaining run provenance are specified below.}
\label{tab:config}
\end{table*}
\paragraph{Run provenance and reporting scope.}
Table~\ref{tab:config} documents the training recipe. Muon updates eligible trainable two-dimensional backbone matrices; a separate AdamW optimizer handles the remaining trainable backbone parameters, and another AdamW optimizer updates the selector. The packed budget is $6\times3{,}072=18{,}432$ positions per distributed training step. This is a per-step capacity, not the cumulative number of training tokens. The recipe targets five epochs by default, but resumption settings can change the final budget. Exact training-data revisions, evaluated checkpoint identifiers and final resumed-run budgets are not recorded in the available run metadata. The saved \texttt{train\_config.yaml}, \texttt{joint\_settings.json}, \texttt{experiment.json} and checkpoint hashes identify any overrides and the checkpoints actually evaluated.

\paragraph{Random seeds.}
The launcher's default initialization seed is 42. Distributed packing and shuffling use sampler seed 0, with epoch $e$ using $0+e$. Saved-checkpoint validation uses seed 4242 inside an RNG-preserving context, so validation restores the surrounding training RNG state. The evaluation commands set both generation and prompt-selection seeds to 42. The launcher default describes the recipe; saved run configurations establish whether a particular training run overrides it.

\paragraph{Numerical aggregation.}
\label{app:aggregation}
The accompanying calculation script recomputes unweighted means from the released benchmark-level entries, retaining all available precision until the final display step. EAGLE-3 speedups use the original throughput values and model-specific AR references; other methods are aggregated from their reported benchmark values. Decimal half-up rounding gives two decimal places for summary entries. Relative changes are calculated from the unrounded computed means, so they need not equal percentages obtained by dividing rounded table entries. Independently supplied sensitivity measurements and profiling results retain their reported precision.

\paragraph{Prefix weighting in the selector loss.}
The selector loss of Eq.~\ref{eq:selloss} is computed on recorded prefixes. A recorded prefix is \emph{eligible} at position $i$ if all earlier recorded tokens equal the cached greedy target tokens and appear in their shortlists:
\begin{equation}
 R_i=\prod_{j=1}^{i-1}\ind[r_j=y_j^*]\,\ind[r_j\in\mathcal C_j],
 \qquad R_1=1.
 \label{eq:reach}
\end{equation}
Positions after the prefix becomes ineligible keep meaningful labels for their own recorded context, but those labels do not describe what verification would accept online. Their weight decays linearly to the floor $\rho=0.25$:
\begin{align}
 \beta(t)&=\clip((t-4000)/4000,0,1),\nonumber\\
 w_i(t)&=1-\beta(t)(1-\rho)(1-R_i),
 \label{eq:weight}
\end{align}
and each covered position contributes $w_i(t)\ell_i$. The selector weight follows $\alpha(t)=0.5\,\clip((t-2000)/2000,0,1)$. The selector loss is computed per worker and its gradients are averaged across workers. Recorded predecessor tokens are embedded directly, even when they are absent from earlier shortlists; no shortlist is modified.

\paragraph{AR branch.}
Of the $n$ anchors sampled per worker, the AR pass uses $c=\min(n,64)$ evenly spaced anchors. Predictor states are projected to the full vocabulary in chunks of 128 positions, and cross-entropy is computed in FP32. The AR loss averages over valid tokens across all workers. The AR pass uses the same document-packing and valid-position restrictions as the selector.

\paragraph{Replay guard.}
Each worker keeps a first-in, first-out buffer of at most 32 detached training examples with their shortlists, adding up to four anchors per step. Before a selector update, a BF16 copy of the selector greedily selects paths on these examples and records the \emph{protected prefix}, the leading choices that equal the cached greedy target. The guard saves the selector parameters $\phi_t$ and optimizer state, then performs one AdamW step, updating the first and second moments and incrementing the optimizer step counter. Let $\Delta$ be the resulting parameter displacement and $g$ the gradient of the shortlist cross-entropy on the protected positions. The guard applies the stabilized projection
\begin{equation}
 \Delta_{\mathrm{stab}}=\Delta-
 \frac{\max(g^\top\Delta,0)}{\max(\|g\|_2^2,\epsilon)}g,
 \label{eq:guard}
\end{equation}
with $\epsilon=10^{-30}$. It tests $\phi_t+s\Delta_{\mathrm{stab}}$ at scales $s\in\{1,\tfrac12,\tfrac14,\tfrac18\}$ in that order and accepts the first scale at which no protected prefix becomes shorter on any worker. An accepted update sets $\phi_{t+1}=\phi_t+s\Delta_{\mathrm{stab}}$ and retains the updated AdamW moments and incremented step counter unchanged. Only the parameter displacement is scaled; moments are neither scaled nor recomputed. Backtracking trials perform no additional optimizer steps and do not increment the counter again. If every trial fails, the previous parameters, first and second moments, and step counter are restored. The learning-rate schedule follows training iterations, including iterations whose selector update is rejected. The guard starts at step 4{,}000, affects the selector only and adds no inference cost. Proposition~5 states what it guarantees.

\section{Lossless Sampling at Temperature One}
\label{app:sampling}
At temperature one, \CPA{} still proposes one deterministic path. Speculative sampling with proposal distribution $q_i$ accepts proposal $\hat y_i$ with probability $\min(1,p_i(\hat y_i)/q_i(\hat y_i))$ and, on rejection, samples from the normalized residual $\max(0,p_i-q_i)$, where $p_i$ is the target distribution after all logit processing, conditioned on the accepted prefix \citep{leviathan2023speculative,chen2023speculative}. For a deterministic proposal, $q_i$ is a point mass on $\hat y_i$. The acceptance probability is then $p_i(\hat y_i)$, and the residual is $p_i$ with $\hat y_i$ removed and renormalized. The probability of emitting $\hat y_i$ is $p_i(\hat y_i)$, and the probability of emitting any other token $v$ is $(1-p_i(\hat y_i))\,p_i(v)/(1-p_i(\hat y_i))=p_i(v)$. Each emitted token therefore follows the target distribution, and induction over positions and rounds gives the target's sequence distribution. The argument does not depend on how the path was selected, as long as the selection does not use the target's random draws.

\section{Additional Results}
\label{app:sensitivity}
\label{app:extended}
\label{app:cost}
\input{tables/extended_results}

Table~\ref{tab:extended} breaks down Table~\ref{tab:components} by benchmark. The full \CPA{} drafter is faster and commits more tokens than both the local-only selector and the drafter trained without the AR loss on every benchmark.

\paragraph{Concurrency.}
Table~\ref{tab:serving-sensitivity} reports output throughput and speedup with 1, 8, 16 and 32 concurrent requests. Each speedup is relative to target-only decoding at the same concurrency.
\input{tables/sensitivity_serving}

\section{Evaluation Benchmarks}
\label{app:data}

Table~\ref{tab:benchmarks} lists the shared dataset versions and splits. EAGLE-3, DFlash, DSpark and \CPA{} are evaluated on the same prompts, using every prompt in each split; for MT-Bench we use the first turn of each question, and for LiveCodeBench the \texttt{release\_v5} problems. MATH-500 is the 500-problem subset of MATH released with PRM800K \citep{huggingfaceh42024math500,openai2023prm800k}. Reported averages give equal weight to each workload. All methods use the target's chat template, thinking mode disabled and the decoding settings in Section~\ref{sec:experiments}.

\begin{table}[H]
\centering\small
\setlength{\tabcolsep}{4pt}
\renewcommand{\arraystretch}{1.15}
\begin{tabular*}{\columnwidth}{@{\extracolsep{\fill}}ll@{}}
\toprule
Benchmark & Hugging Face dataset; configuration / split\\
\midrule
GSM8K & {\scriptsize\texttt{openai/gsm8k}; \texttt{main} / \texttt{test}}\\
MATH-500 & {\scriptsize\texttt{HuggingFaceH4/MATH-500}; \texttt{test}}\\
AIME 2025 & {\scriptsize\texttt{MathArena/aime\_2025}; \texttt{train}}\\
HumanEval & {\scriptsize\texttt{openai/openai\_humaneval}; \texttt{test}}\\
MBPP & {\scriptsize\texttt{google-research-datasets/mbpp};}\\
 & {\scriptsize\texttt{sanitized} / \texttt{test}}\\
LiveCodeBench & {\scriptsize\texttt{livecodebench/code\_generation\_lite};}\\
 & {\scriptsize\texttt{release\_v5}}\\
MT-Bench & {\scriptsize\texttt{HuggingFaceH4/mt\_bench\_prompts};}\\
 & {\scriptsize\texttt{train}, first turn}\\
Alpaca & {\scriptsize\texttt{tatsu-lab/alpaca}; \texttt{train}}\\
\bottomrule
\end{tabular*}
\caption{Benchmark sources shared by EAGLE-3, DFlash, DSpark and \CPA{}. All methods use every prompt in the listed split with the same evaluation settings.}
\label{tab:benchmarks}
\end{table}

%% file: sections/appendix_theory.tex
\section{Formal Properties and Numerical Scope}
\label{app:theory}

The statements below concern the one-layer mathematical selector on a fixed block. The fixed-pair compilation and walk-cost guarantees do not extend to the stacked \texttt{local\_prefix\_v2} implementation, whose second-layer query depends on the selected prefix through its first layer (Appendix~\ref{app:stacked}). Let $m$ be the number of proposed positions, $C_i=\{c_{i,1},\ldots,c_{i,K}\}$, and $C_0=\{a\}$ for the committed anchor. Set $k_0=1$. All block features are held fixed during selection. We suppress dependence on backbone parameters $\theta$ and selector parameters $\phi$.

\paragraph{Proposition 1: Exact table compilation.}
For $j\leq i-2$, suppose the positive pair weight $A_{ij}^{h}[k,l]$ and weighted readout $B_{ij}^{h}[k,l]=A_{ij}^{h}[k,l]d_{ij}^{h}(c_{i,k},c_{j,l})$ depend only on fixed block features and the indicated token pair. Maintain a separate pair of accumulators for every destination, candidate, and head:
\begin{align}
 N_i^h[k]&=\sum_{j=0}^{i-2}B_{ij}^h[k,k_j],\\
 D_i^h[k]&=1+\sum_{j=0}^{i-2}A_{ij}^h[k,k_j].
\end{align}
Empty sums are zero. Initialize $N$ to zero and $D$ to one; process the anchor column for every $i\geq2$. Whenever position $j\geq1$ selects $k_j$, add $B_{ij}^h[:,k_j]$ and $A_{ij}^h[:,k_j]$ only to destinations $i\geq j+2$. Then
\begin{align}
 S_i(c_{i,k})={}&u_i(c_{i,k})
 +L_i(c_{i,k},c_{i-1,k_{i-1}})\nonumber\\
 &+\sum_{h=1}^{H}\omega_h\frac{N_i^h[k]}{D_i^h[k]}
\end{align}
equals direct evaluation of the documented local and older-history terms in exact arithmetic.

\emph{Proof.} After processing sources $0,\ldots,j$, each destination accumulator contains its initialization plus exactly the selected table columns with source index at most $\min(j,i-2)$. This invariant holds initially and is preserved by the stated additions. Immediately before choosing position $i$, its included sources are therefore exactly $0,\ldots,i-2$. Substituting the table definitions gives the direct attention numerator and denominator. The immediate predecessor appears only in $L_i$. Thus scores agree; a common deterministic tie rule gives the same choice at each step, by induction. In particular, the history term is zero at position one and uses only the anchor at position two.\hfill$\square$

\paragraph{Packed table size.}
An implementation storing only needed entries uses
\begin{align}
 n_L &= K+(m-1)K^2,\\
 n_A=n_B &= HP,\\
 P&=(m-1)K+\tfrac{(m-1)(m-2)}{2}K^2.
 \label{eq:paircount}
\end{align}
For $m=15$, $K=16$, and $H=4$, these are $3{,}600$ local entries and $94{,}080$ entries in each history table. The combined $191{,}760$ scalars occupy approximately $749.1$ KiB in FP32 or $374.5$ KiB in BF16. These are packed mathematical table sizes, excluding projections, embeddings, accumulators, padding, and execution workspaces; they are not measured device-memory allocations. Table storage is $O(Hm^2K^2)$, while incremental history gathering and accumulation require $O(Hm^2K)$ work with $m$ sequential choices. The exact projection and table-construction costs depend on the specified pair-feature parameterization.

\paragraph{Proposition 2: Shortlist prefix ceiling.}
For a fixed verified context including anchor $a$, let $z_{1:m}$ be the target's true greedy continuation, and let $C_{1:m}$ be the fixed candidate sets produced for that block. Define
\begin{equation}
 R_C=\min\bigl(\{i:z_i\notin C_i\}\cup\{m+1\}\bigr)-1.
\end{equation}
Any selector constrained to $C_i$ has at most $R_C$ accepted draft tokens before its first mismatch.

\emph{Proof.} If $R_C<m$, then $z_{R_C+1}\notin C_{R_C+1}$, so no allowed proposal can match the target at that position after matching all earlier ones. The case $R_C=m$ is immediate. For any distribution of blocks, the tail-sum identity therefore gives
\begin{equation}
 \mathbb{E}[A]\leq\mathbb{E}[R_C]
 =\sum_{\ell=1}^{m}\Pr\!\left(\bigcap_{i=1}^{\ell}\{z_i\in C_i\}\right).
\end{equation}
\hfill$\square$

For sequences ending before $m$ positions, both quantities are truncated at the same EOS or remaining-output boundary. The joint coverage probabilities need not factor into marginal hit rates. Moreover, cached target labels conditioned on a different recorded continuation cannot substitute for $z_{1:m}$ after that continuation diverges from the true greedy rollout.

\paragraph{Proposition 3: Unary preservation at initialization.}
Suppose both correction paths are identically zero, all intermediate quantities are finite, and the full-vocabulary unary winner $v_i^\star$ is retained as candidate slot one. If candidate score ties favor slot one, initialized selection chooses $v_i^\star$ at every position.

\emph{Proof.} Every candidate score equals its unary logit, and no candidate exceeds the full-vocabulary maximum. Slot one achieves that maximum and wins any tie.\hfill$\square$

This statement requires compatible unary and selector tie rules. An unspecified top-$K$ ordering is insufficient when BF16 logits tie. It establishes initialization behavior only.

\paragraph{Proposition 4: Greedy verification preserves target output.}
Assume a deterministic target decision rule, including identical logits processors and tie handling, and verification logits that equal the corresponding target causal conditionals. Starting from a committed context, accept the longest proposed prefix matching those target decisions. If a mismatch occurs, append the target decision at that position; if the full block matches, append the target's next decision as a bonus. Stop immediately at EOS and obey the remaining output budget. The committed output is identical to serial target greedy decoding.

\emph{Proof.} The first verified position is evaluated under the committed context. Each accepted match extends that context by exactly the serial target decision, so induction establishes equality at every accepted position. The correction, or the bonus after a fully accepted block, is evaluated under the resulting matching prefix and is therefore also the serial target decision. Repeating this argument across cycles proves equality until stopping. The anchor is already part of the committed context and is not emitted a second time.\hfill$\square$

The proposition covers greedy decoding. An exact stochastic extension requires a separately specified proposal distribution and acceptance/correction sampler. In finite precision, batched verification and serial execution must be checked for target-decision parity; identical mathematical models alone do not ensure bitwise equal logits across kernels.

\paragraph{Proposition 5: Scope of the stabilized guard correction.}
Let $g$ be the protection-loss gradient with respect to $\phi$, let $\Delta$ be the proposed AdamW displacement, and let $\epsilon>0$. The documented correction is
\begin{equation}
 \Delta_{\mathrm{stab}}=\Delta-
 \frac{\max(g^\top\Delta,0)}{\max(\lVert g\rVert_2^2,\epsilon)}g.
\end{equation}
If $g^\top\Delta\leq0$, the displacement is unchanged. If $a=g^\top\Delta>0$ and $n=\lVert g\rVert_2^2$, then
\begin{equation}
 g^\top\Delta_{\mathrm{stab}}
 =a\left(1-\frac{n}{\max(n,\epsilon)}\right).
\end{equation}
Thus the first-order increasing component is removed when $n\geq\epsilon$, but remains positive when $0<n<\epsilon$. This follows directly by taking the inner product with $g$. Consequently the stabilized formula is not universally the Euclidean projection onto the non-increasing halfspace. The exact nonzero-gradient projection uses $n$ in the denominator; any small-norm alternative must be specified as a distinct implementation choice.

For fixed detached replay features, candidate sets, and deterministic BF16 evaluation, the subsequent discrete test supplies a narrower guarantee: a trial accepted only when every replay example's protected prefix length is at least its pre-update value preserves those tested lengths. All distributed ranks must satisfy the condition. The implementation takes one AdamW step before projecting and backtracking its parameter displacement. An accepted scaled trial retains that step's updated moments and incremented counter unchanged; trials do not execute additional optimizer steps. Full rejection restores the previous selector parameters, moments and counter. The learning-rate schedule continues to follow training iterations (Appendix~\ref{app:guard}). This statement does not extend to changed backbone features, new shortlists, unseen inputs, online acceptance, or latency.

\paragraph{Finite-precision compilation.}
Proposition~1 establishes algebraic equality, not bitwise equality after changed summation order, casts, or fused kernels. At a given step, suppose compiled and direct scores differ by at most $\delta$ per candidate under the same preceding choices. A direct top-two margin strictly greater than $2\delta$ preserves that step's winner: the winner can decrease by at most $\delta$, while any competitor increases by at most $\delta$. Applying this argument inductively preserves the entire selected path if the condition holds at every reached step. Near ties require explicit token-level parity checks, including the deployed BF16 parameter values and captured-graph execution. Warmup parity is a finite implementation check, not a proof for all future blocks.

%% file: tables/extended_results.tex
\begin{table*}[t]
\centering
\caption{\textbf{Per-benchmark ablation.} Greedy Qwen3-4B; metrics and aggregation as in Table~\ref{tab:main}. Parallel argmax and Local only switch off the selector or its history term at inference on the \CPA{} checkpoint; No AR loss is a drafter trained with $\lambda_{\rm AR}=0$. Means and derived percentages use the supplied workload values before rounding the aggregates.}
\label{tab:extended}
\begingroup
\fontsize{8}{9.4}\selectfont
\setlength{\tabcolsep}{1.55pt}
\renewcommand{\arraystretch}{1.13}
\begin{tabular*}{\textwidth}{@{\extracolsep{\fill}}l*{18}{c}@{}}
\toprule
& \multicolumn{6}{c}{\textsc{Math}}
& \multicolumn{6}{c}{\textsc{Code}}
& \multicolumn{4}{c}{\textsc{Chat}}
& \multicolumn{2}{c}{\textsc{Overall}}\\
\cmidrule(lr){2-7}
\cmidrule(lr){8-13}
\cmidrule(lr){14-17}
\cmidrule(lr){18-19}
Method
& \multicolumn{2}{c}{GSM8K}
& \multicolumn{2}{c}{MATH-500}
& \multicolumn{2}{c}{AIME25}
& \multicolumn{2}{c}{HumanEval}
& \multicolumn{2}{c}{MBPP}
& \multicolumn{2}{c}{LCB}
& \multicolumn{2}{c}{MT-Bench}
& \multicolumn{2}{c}{Alpaca}
& \multicolumn{2}{c}{Avg.}\\
\cmidrule(lr){2-3}
\cmidrule(lr){4-5}
\cmidrule(lr){6-7}
\cmidrule(lr){8-9}
\cmidrule(lr){10-11}
\cmidrule(lr){12-13}
\cmidrule(lr){14-15}
\cmidrule(lr){16-17}
\cmidrule(lr){18-19}
& $S$ & $\tau$
& $S$ & $\tau$
& $S$ & $\tau$
& $S$ & $\tau$
& $S$ & $\tau$
& $S$ & $\tau$
& $S$ & $\tau$
& $S$ & $\tau$
& $S$ & $\tau$\\
\midrule
Parallel argmax
& $4.07\times$ & 8.64
& $4.44\times$ & 8.69
& $3.69\times$ & 6.98
& $3.33\times$ & 6.48
& $2.85\times$ & 5.73
& $3.29\times$ & 6.31
& $1.59\times$ & 3.80
& $1.78\times$ & 3.48
& $3.13\times$ & 6.26\\
Local only
& $4.11\times$ & 9.24
& $4.49\times$ & 9.22
& $3.77\times$ & 7.60
& $3.40\times$ & 6.98
& $3.00\times$ & 6.36
& $3.23\times$ & 6.58
& $1.73\times$ & 4.46
& $1.84\times$ & 3.86
& $3.20\times$ & 6.79\\
No AR loss
& $4.29\times$ & 9.56
& $4.58\times$ & 9.33
& $3.90\times$ & 7.81
& $3.43\times$ & 6.94
& $3.08\times$ & 6.44
& $3.24\times$ & 6.50
& $1.91\times$ & 4.97
& $1.84\times$ & 3.85
& $3.28\times$ & 6.93\\
\textbf{\CPA{}}
& $4.38\times$ & 9.85
& $4.68\times$ & 9.62
& $3.99\times$ & 8.05
& $3.49\times$ & 7.17
& $3.12\times$ & 6.61
& $3.29\times$ & 6.71
& $1.98\times$ & 5.12
& $1.88\times$ & 3.95
& $3.35\times$ & 7.14\\
\bottomrule
\end{tabular*}
\endgroup
\end{table*}

%% file: tables/sensitivity_serving.tex
\begin{table}[H]
\centering
\caption{\textbf{Throughput under concurrent requests.} Qwen3-4B, greedy, all evaluation prompts across the eight benchmarks, at most 2{,}048 new tokens, one Ascend 910B2 NPU. Each cell gives output tokens/s and, below it, speedup over target-only decoding at the same concurrency.}
\label{tab:serving-sensitivity}
\fontsize{8.5}{10.3}\selectfont
\renewcommand{\arraystretch}{1.4}
\setlength{\tabcolsep}{3pt}
\begin{tabular*}{\columnwidth}{@{\extracolsep{\fill}}rrrrr@{}}
\toprule
Concurrency & Target AR & DFlash & DSpark & \textbf{CPA}\\
\midrule
1 & \shortstack{47.83\\($1.00\times$)} & \shortstack{143.49\\($3.00\times$)} & \shortstack{149.23\\($3.12\times$)} & \shortstack{\textbf{160.23}\\($\mathbf{3.35}\times$)}\\
8 & \shortstack{248.17\\($1.00\times$)} & \shortstack{564.88\\($2.28\times$)} & \shortstack{586.26\\($2.36\times$)} & \shortstack{\textbf{631.22}\\($\mathbf{2.54}\times$)}\\
16 & \shortstack{452.78\\($1.00\times$)} & \shortstack{805.54\\($1.78\times$)} & \shortstack{836.60\\($1.85\times$)} & \shortstack{\textbf{895.11}\\($\mathbf{1.98}\times$)}\\
32 & \shortstack{649.57\\($1.00\times$)} & \shortstack{883.01\\($1.36\times$)} & \shortstack{921.46\\($1.42\times$)} & \shortstack{\textbf{986.39}\\($\mathbf{1.52}\times$)}\\
\bottomrule
\end{tabular*}
\end{table}